\documentclass[journal,comsoc]{IEEEtran}

\IEEEoverridecommandlockouts

\pdfoutput=1

\usepackage{graphicx}
\usepackage{algorithm}
\usepackage{algorithmic}
\usepackage{subcaption}

\usepackage[bottom]{footmisc}
\usepackage{cite}
\usepackage{url}
\usepackage{mdwmath}     
\usepackage{mdwtab}
\usepackage{amsfonts,amsmath,color,amssymb,amsxtra,amsbsy,floatflt}
\usepackage{indentfirst} 
\usepackage{tabularx}
\usepackage{booktabs}
\usepackage{multirow}
\usepackage{nth}
\usepackage{comment}
\usepackage{epstopdf}
\usepackage{bbold}
\usepackage{hhline}
\usepackage{dblfloatfix}
\usepackage[table]{xcolor}      

\usepackage{mathtools}

\newcommand{\vv}[1]{\boldsymbol{#1}}
\newcommand{\vvi} [2]{ \mathbf{#1}^{(#2)} }

\newcommand{\change}[1]{{\color{black}{#1}}}

\newcommand{\name}{ARENA}

\newcommand{\st}[1]{\text{s.t.}~#1}

\newcommand{\norm}[1]{\left\lVert#1\right\rVert}

\begin{document}
\graphicspath{{./Figures/}}
\title{ARENA: A Data-driven Radio Access \\ Networks Analysis of Football Events}

\author{Lanfranco~Zanzi,~\IEEEmembership{Associate Member,~IEEE,}
	    Vincenzo~Sciancalepore,~\IEEEmembership{Senior Member,~IEEE,}
   	    Andres~Garcia-Saavedra,
   	    Xavier~Costa-P\'erez,~\IEEEmembership{Senior Member,~IEEE,}
   	    Georgios~Agapiou,
        and~Hans~D.~Schotten,~\IEEEmembership{Member,~IEEE}
    \thanks{\textit{L. Zanzi is with NEC Laboratories Europe GmbH, Heidelberg, Germany, and Technische Universit\"at Kaiserslautern, Kaiserslautern, Germany. Email: lanfranco.zanzi@neclab.eu.}}%
	\thanks{\textit{V. Sciancalepore, A. Garcia-Saavedra are with NEC Laboratories Europe GmbH., Heidelberg, Germany. Emails: \{vincenzo.sciancalepore, andres.garcia.saavedra\}@neclab.eu.}}%
	\thanks{\textit{X. Costa-P\'erez is with NEC Laboratories Europe GmbH, Heidelberg, Germany, and i2CAT Foundation and ICREA, Barcelona, Spain. Email: xavier.costa@neclab.eu.}}%
	\thanks{\textit{Hans D. Schotten is with Technische Universit\"at Kaiserslautern, Kaiserslautern, Germany. Email: schotten@eit.uni-kl.de.}}%
	\thanks{\textit{G. Agapiou is with OTE, Athens, Greece. Email: gagapiou@oteresearch.gr}}%
}

\maketitle
\thispagestyle{empty}	

\begin{abstract}
Mass events represent one of the most challenging scenarios for mobile networks because, although their date and time are usually known in advance, the actual demand for resources is difficult to predict due to its dependency on many different factors. Based on data provided by a major European carrier during mass events in a football stadium comprising up to $30.000$ people, $16$ base station sectors and $1$~Km$^2$ area, we performed a data-driven analysis of the radio access network infrastructure dynamics during such events.

Given the insights obtained from the analysis, we developed \name{}, a model-free deep learning Radio Access Network (RAN) capacity forecasting solution that, taking as input past network monitoring data and events context information, provides guidance to mobile operators on the expected RAN capacity needed during a future event. Our results, validated against real events contained in the dataset, illustrate the effectiveness of our proposed solution.


\end{abstract}


\section{Introduction}

Mass  events such as sport events (e.g., football games), religious events (e.g., holy pilgrimages), political events (e.g., demonstrations) or entertainment events (e.g., concerts) are particularly challenging for mobile operators \emph{despite being planned} with months or weeks ahead in most cases~\cite{Abdallah_2016}. Indeed, operators know \emph{when} a demand surge spawning from such events is coming but they are unable to adapt timely and appropriately~\cite{7997091}.

\change{Traditional delay tolerant~\cite{Delay-Tolerant} and information centric~\cite{Information-Centric} approaches offer methods to smooth the traffic volumes during congestion time, e.g., postponing the transmission of delay tolerant traffic outside the busy time windows, or reducing the traffic flowing in the network by a proactive content placement at the edge of the network. However, due to generally scarce radio access resource availability, none of those approaches would avoid mobile traffic congestion in the event premises. To increase the spectrum availability, current approaches imply the on-demand deployment of, e.g., nomadic cells such as Cells on Wheels (CoWs) or Cells on Light Trucks (CoLTs), or fixed small (micro, pico, femto) cells for surge offloading. All these options are rather rudimentary and, needless to say, overly costly.}
Network slicing~\cite{Slicing2018CoNEXT}, which will allow an operator to request further slices of dormant resources (even from other operators) in a more flexible manner, can certainly reduce the cost tied to the hunt for capacity during these events. However, albeit planned ahead in time, the amount of additional resources required  to cope with the demand during these situations is largely unpredictable. 

Indeed, the amount of network load resulting from these events \emph{depends on contextual features} such as the type of event---different events foster different mobile applications, such as real-time video streaming during concerts; and consumption patterns, such as data avalanches occurring during the breaks of a football match---or the ability of the event to attract attendance, such as the ranking of the matching teams in a football competition. However, although this contextual information is available, the model capturing the relationship between mobile traffic demand and the specific context of a given event  \emph{is inherently hard to build because mass events are rare and each is different in nature from one another}. 

In this paper, we advocate for the use of model-free deep learning techniques to take up on the challenge. Specifically, we first analyze data obtained from a major operator regarding the cellular coverage in the stadium of an important football team in Greece. It is well understood nowadays that the use of radio spectrum through mobile systems is closely related to human activity~\cite{Slicing_Infocom19}. In this way, our analysis not only makes evident the aforementioned challenges, but it also sheds light on the behavioral dynamics of sport fans when attending major football events. Then, we formulate a Bandit Convex Optimization (BCO) model to formalize the problem we are addressing. Our model formalizes some of the aspects that make capacity forecasting during mass events complex, that is, the unknown relationship between user traffic patterns, spectrum usage and the context around the event. Then, to address this problem we design a deep neural network model to estimate the amount of extra spectrum resources required during the event to preserve the quality of service experienced by end users during regular days.

To summarize, the main contributions are the following:
\begin{itemize}
    \item We provide a quantitative and qualitative analysis of the network statistics during football events taken from a real deployment;
    \item We formulate our problem by means of a Bandit Convex Optimization (BCO) model;
    \item We evaluate the performance of different state-of-the-art forecasting models while dealing with real-data;
    \item We propose a deep learning architecture, namely \name{} that takes as input monitoring metrics from Radio Access Network (RAN) devices as well as other contextual information to assess the extra spectrum resources required during the event;
    \item We validate the model and finally quantify the expected amount of resources to be proactively deployed in the stadium area to meet the target Quality of Service (QoS).
\end{itemize}

The rest of the paper is structured as follows.
Section~\ref{sec:related} collects the state-of-the-art and related works in the field.
Section~\ref{sec:scenario} discusses the scenario considered in this work.
Section~\ref{sec:analysis} presents an overview of the dataset of networking measurements and features therein contained, providing insight on the public event dynamics from the mobile network perspective, with focus on the quality of service experienced by the end users.
In Section~\ref{sec:model} we model the network dynamics in the stadium by means of bandit optimization theory and propose in Section~\ref{sec:ML} a machine-learning (ML) solution, namely \name{}, to capture the relationship between user activity and network utilization.
Section~\ref{sec:evaluation} evaluates \name{}'s capabilities to forecast future network dynamics and estimate the additional amount of radio resources to improve the end-user QoS.
Finally, Section~\ref{sec:conclusions} concludes the paper.

\paragraph*{Privacy issues}
Our research activity does not violate user's privacy rights. The dataset contains only high-level aggregated and anonymous information. 

\section{Related Work}
\label{sec:related}

Mass events (sports, in particular) are gaining substantial attention for analysis in the last few years~\cite{Morrison2009, Superbowl2013, CastagnoWoWMoM17}. This is certainly a projection of the underlying social attraction to novel applications (and the business therein) that improve the experience for attendees and participants, e.g., virtual/augmented reality (VR/AR) technologies, in the upcoming 5G reign. 

\change{An accurate characterization of mobile network channel statistics is of paramount importance to assess the network capacity boundaries. However, as highlighted in~\cite{MONROE}, stable performances in mobile networks are much more complex to be achieved due to highly variable wireless channel statistics that affect the final service provisioning.}
The authors of \cite{Morrison2009} present a detailed analysis of spectator behavior with the goal of designing novel mobile applications for stadium-based sporting events. In their work, they exploit Bluetooth and GPS traces to derive information such as density, location and even travel speeds of crowds of people attending the UEFA world cup final that took place at the City of Manchester Stadium in May $2008$. This early study hints at the importance of capturing contextual information to manage mobile services during mass events. 
A thorough analysis of mobile network performance during the 2013 Superbowl is presented in~\cite{Superbowl2013}. The authors conclude that the use of multicast cellular coverage in combination with content caching is paramount to solve the congestion of the uplink LTE channels during super-sized events.
The authors of~\cite{CastagnoWoWMoM17} analytically investigate the root causes of cellular network performance drops during crowded events, identifying the random access procedure as the main cause for QoE degradation and suggesting to leverage on device-to-device communication to cluster network access requests as to alleviate the problem. Similarly,~\cite{Mandalari_Mobicom18} provides an in-depth analysis of the subscribers’ quality of experience focusing on roaming scenarios.

Mobile traffic forecasting has fostered substantial research effort in its own merit.
In~\cite{Exploring_Human_Mobility,Cellular_Traffic_Metropolis,Crowd_Flow_Prediction, Xu_Mobile_Traffic_Patterns} the authors explore mobile traces collected from metropolitan areas to investigate and understand human mobility.
In~\cite{gramaglia_data_orange, gramaglia_data_orange2} the authors focus on mobile resource utilization of individual services at a national scale. This work sheds the lights on interesting macroscopic properties and provides informative insights of todays' mobile networks, including spatio-temporal traffic patterns as well as the relationship between data volumes and urbanization levels. 
Moreover, several machine learning mobile traffic prediction solutions have been proposed in the literature.
In~\cite{Wang_Infocom18}, the authors use deep learning techniques for spatio-temporal modeling and prediction in cellular networks.
The authors of~\cite{Slicing2018CoNEXT} apply forecasting of mobile network resource utilization to investigate the network slicing concept in 5G systems. They design a dynamic resource allocation framework accounting for all the network domains and validate their proposal through exhaustive simulation campaigns. Deep models for long-term traffic prediction are also provided in \cite{Zhang2018,Cellular_Traffic_Metropolis,Urban_Traffic_Prediction}. Of particular interest from these works is how deep learning models substantially outperform traditional predictive methods. 
Finally, \cite{DeepCogInfocom19} proposes an encoder-decoder neural-network structure to forecast traffic demands on specific services (or network slices) and solve the trade-off between capacity over-dimensioning and service requirements guarantees. While this work on traffic forecasting provides insightful understanding on the dynamics of regular-user regular-day when interacting with mobile services, the model capturing such behavior is not valid during rare but massively crowded events. 

In the sequel, we will first analyze the case of an important football stadium in Greece and the mobile coverage provided by a major mobile operator during a regular football season. A key observation of our analysis is that the relationship between active users and radio resources during sport events substantially differ from that of a regular day and so it requires special consideration to that given in \cite{gramaglia_data_orange, gramaglia_data_orange2, Wang_Infocom18, Slicing2018CoNEXT, DeepCogInfocom19}.

\begin{figure}[t!]
\centering
\includegraphics[width=\columnwidth, clip, trim = 0cm 0cm 0cm 0cm]{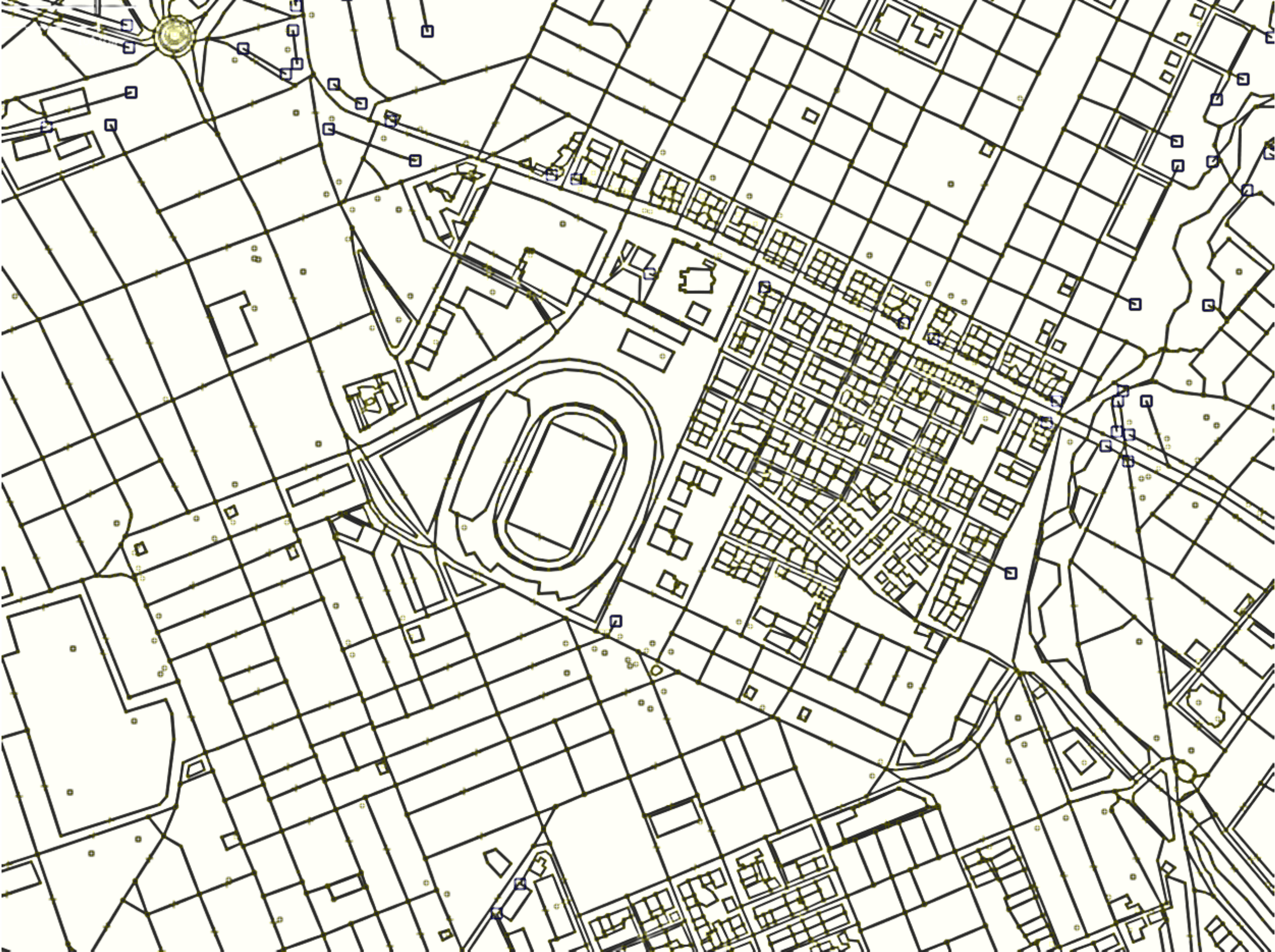}
\caption{Overview of the stadium area.}
\label{fig:stadium}
\end{figure}

\section{Preliminary considerations}
\label{sec:scenario}


Despite frustrating, the poor network performance achieved in crowded initiatives like concerts or sport matches is commonly accepted by public and professionals attending such events who, after several initial unsuccessful trials, often postpone the upload (or download) of their photos and videos on social media due to network congestion. Similar poor conditions are mostly registered during an event in case of voice calls with a clear dropping rate increase~\cite{performance_crowded_events}.
The problem is well known and accounts for two separate root causes deriving from the need of widely guaranteeing access to the network, i.e., congestion in the radio access network (RAN) during the random access procedure (uplink traffic) and limited resource availability against a huge data demand surge (downlink traffic).

\subsection{Standardization procedures}
During the initial access, the User Equipment (UE) scans the system broadcast information to obtain the main system configuration parameters and synchronize with the network before attempting to attach through the physical random access channel (PRACH).
The LTE standard defines two random access procedures: a contention-based and a contention-free. The former is mainly used for initial access, Radio Resource Control (RRC) layer connection (re-)establishment and for uplink data transmission in non-synchronized states or in absence of scheduled resource availability. The latter, less common, is mostly implemented to allow fast handover procedures. 
The RRC protocol of LTE system includes, among the others, connection establishment and connection release, system information broadcast and radio bearer setup and reconfiguration~\cite{TS36.331}. In LTE, the base stations or eNBs centralize most of the RRC functionality as well as the resource management and packet scheduling. The set of activities performed by the eNB varies according to the state of the radio connection---which can be either active (Connected) or not active (Idle)---between the UE and the network. In the $\texttt{RRC}_\texttt{IDLE}$ state, the UE passively monitors the system information broadcast by the network without the need to send monitoring reports or mobility updates. Conversely, UEs move into $\texttt{RRC}_\texttt{CONNECTED}$ state when performing a call or transmitting data, which increase the monitoring information reports sent to the radio access node in order to keep the session active.
During the contention-based procedure, the UE randomly selects one of the $64 - N_{\textit{cf}}$ orthogonal preamble signatures and the next available subframe for PRACH transmission, where $N_{\textit{cf}}$ is the number of preamble signatures allocated for the contention-free procedure (its value can change according to traffic load in the system).
After sending the attachment request, the UE monitors the physical downlink control channel (PDCCH) and schedules a timer. In case of simultaneous RACH requests, the eNB uses the preamble sequence to differentiate among users. However, specially in crowded areas, different requests may collide. In this case, the eNB will not be able to decode the requests thereby triggering an exponential backoff procedure in the UE which delays the next access attempt. If no other user selects the same preamble sequence, the eNB is able to decode the request and reply with a Random Access Response (RAR) message through the PDCCH. At this point, the UE sends a RRC connection request message including a temporary identifier and the establishment cause. If accepted by the network, the access procedure terminates, and the UE moves into $\texttt{RRC}_\texttt{CONNECTED}$ state so that it is allowed to use the data communication services of the system.

Several optimization steps can be adopted to increase the performance of the RACH procedure. Typical deployments assume collision probability between UEs in the order of $1\%$ with periodic random access occasions distributed every $10$ ms. This setup translates into the possibility to handle an offered load of $128$ attempts/second over $10$MHz bandwidth~\cite{LTEBook}. Clearly, such a setup leads to low performance when dealing with crowded events. Typical approaches to solve this situation involve the reduction of the access occasion cycle duration and the increase of access opportunities within one frame, however this negatively impacts on the PRACH overhead finally reducing the scheduling opportunities for user data transmissions.

\subsection{Technical challenge}
The second issue, data demand surges, is more intuitive and involves reduced rate of transmission opportunities  (delay) and small chunks of resources scheduled per user within the Physical Downlink Shared Channel (PDSCH) (low throughput) due to a high number of users in $\texttt{RRC}_\texttt{CONNECTED}$ state. 

Our work focuses on the latter issue. A simple approach to solve it is to smooth down traffic patterns by delaying load opportunistically to time instances with a less-congested network~\cite{Superbowl2013, Delay-Tolerant}. However, this approach is only acceptable for delay-tolerant applications and hardly fits with the real-time nature of most use cases motivating 5G applications. The only solution upon such context is to increase the capacity of the system by deploying nomadic cells such as Cells on Wheels (CoWs) or Cells on Light Trucks (CoLTs), or offload traffic to fixed small cells, which effectively increases the capillarity of the network and hence the density of radio access points.

Unfortunately, this approach is overly expensive and slow, incapable of adapting to different traffic patterns occurring during events of different nature or events with different number of attendees. Certainly, network slicing opens the door for novel ways of increasing the capacity to the network opportunistically in a more agile manner (e.g., leasing additional spectrum through a network slice from neighbouring operators)---in the matter of few hours or even minutes. Still, the human relationship with mobile applications and the number of attendees substantially differ from one event to another. Summing up the fact that such massive events are rare (though planned), makes forecasting capacity requirements particularly daunting. 

\setcounter{figure}{2}
\begin{figure*}[!b]
    \centering
    \includegraphics[width=2\columnwidth, clip, trim = 0cm 0cm 0cm 0cm]{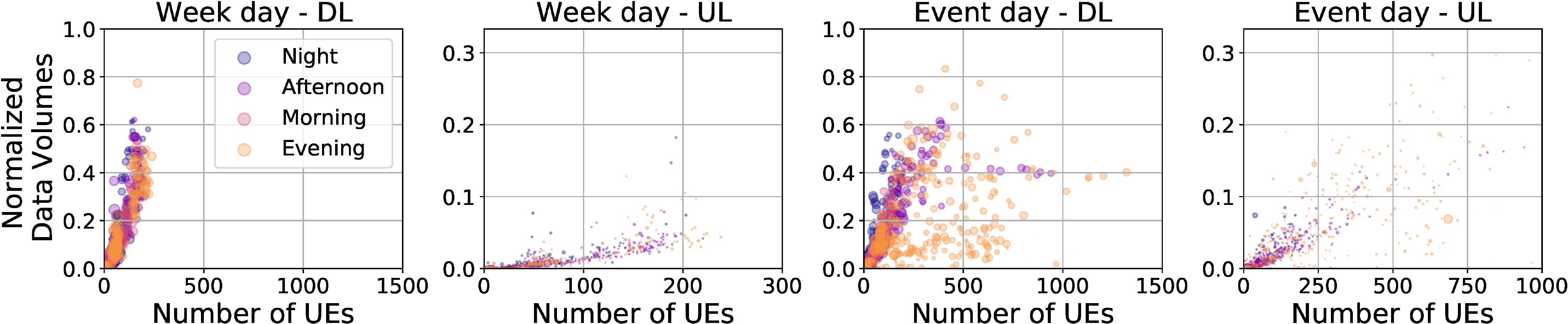}
    \caption{\change{Normalized data volume distribution as a function of the number of active users ($\texttt{RRC}_\texttt{CONNECTED}$) for downlink (DL) and uplink (UL) and different periods of the day (morning, afternoon, evening, night).}}
	\label{fig:scatter_volumes}
\end{figure*}

\section{Data Analysis}
\label{sec:analysis}

\setcounter{figure}{1}
\begin{figure}[!t]
\centering
\includegraphics[width=\columnwidth, clip, trim = 0cm 0cm 0cm 0cm]{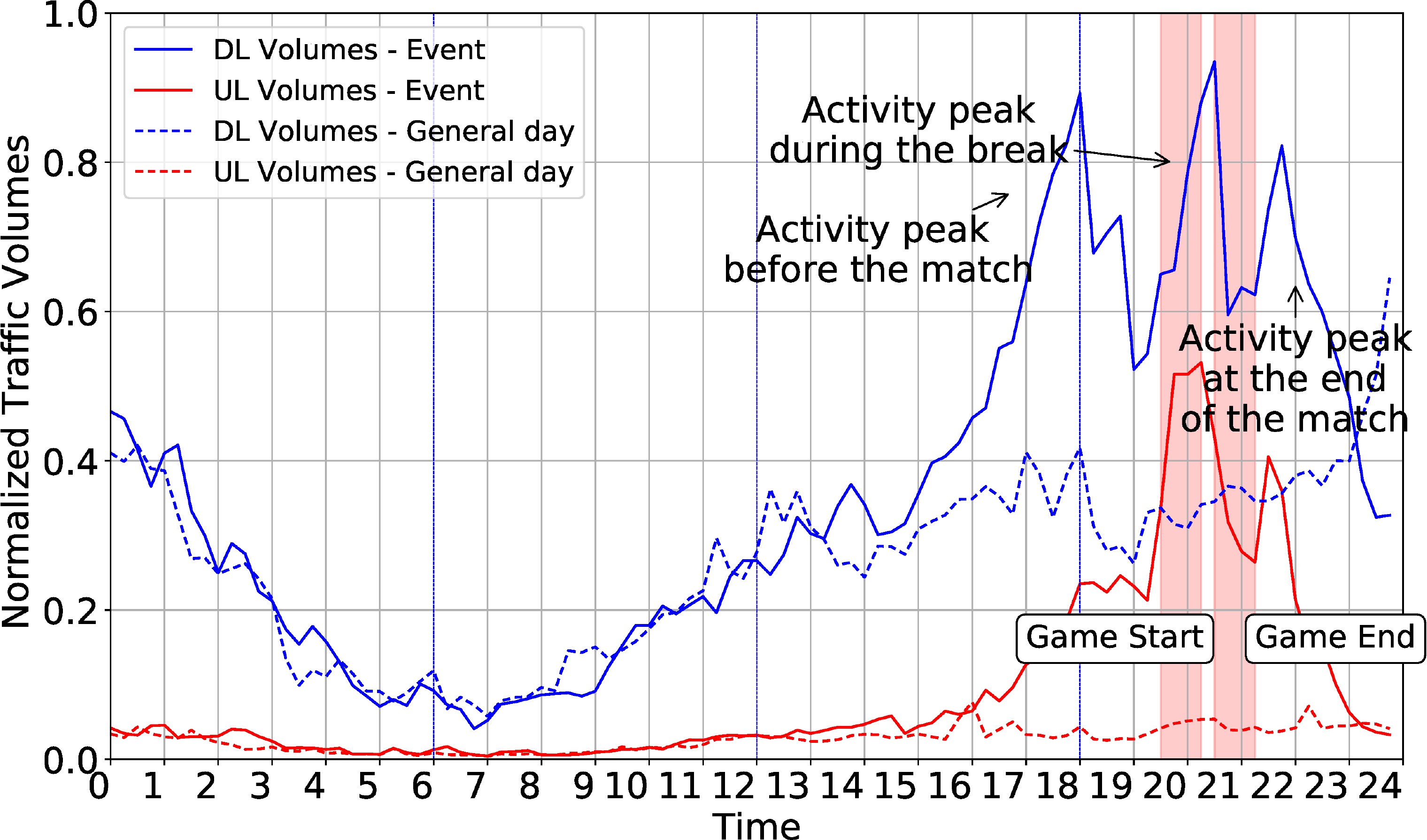}
\caption{Temporal evolution of the aggregated downlink (DL) and uplink (UL) traffic volume in the stadium area during a regular day (dashed line) and during the day of a large event (straight line).}
\label{fig:traffic_volumes}
\end{figure}

In this section we present a data-driven analysis of sport events. Our dataset consists of weeks of monitoring data including large sport events taken place in a football stadium during a regular season under the coverage of $6$ LTE eNBs, subdivided into $16$ sectors, from a major operator in Greece. The set of eNBs covers an area of approximately $1$~Km$^2$. Such concentration of radio access points is required to accommodate the traffic peaks generated by the almost $30000$ people hosted in the stadium during sport events. For the same reason, the base stations are equipped with directive arrays of antennas and support Carrier Aggregation (CA). As shown in Fig.~\ref{fig:stadium}, the surrounding district includes both residential neighbourhoods and vehicular streets, which highly characterize the spatio-temporal behaviour of the traces as later detailed in Section~\ref{subsec:TemporalDistribution}. We do not disclose the location of the eNBs due to privacy matters.
Our goal is to characterize the relationship between active users during the events and network usage in contrast to regular days. Therefore, we focus our analysis on the average throughput per active user experienced by the event attendees in the stadium. Throughout the rest of the paper, we will consider it as the key performance indicator (KPI) of the perceived quality of service (QoS) experienced by the end-users.
To ease our analysis, we focus on a particular sport match, taken place in February $10^{th}$, $2019$, which gathered $25097$ attendees in the stadium as stated by official notes from the organizers.\footnote{While our analysis is based on a single event, we have exhaustively evaluated other $10$ different matches finding no relevant performance differences, which makes our approach generalizable.} The data collection exploits local information registered by the LTE eNBs for monitoring purposes, and include both averaged and aggregated features with a time granularity of $15$ minutes. This ensures that the subscribers' privacy rights are preserved. 


\setcounter{figure}{4}
\begin{figure*}[b!]
    \centering
          \includegraphics[trim = 0cm 0cm 0cm 0cm,clip, width=2\columnwidth]{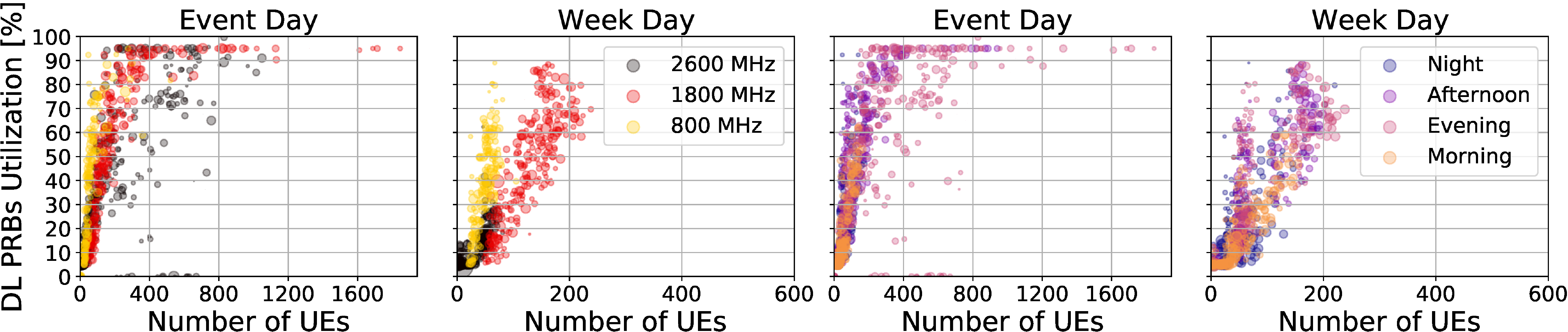}
    \caption{ \change{Downlink PRB utilization as a function of the number of active users during the day of the event and during a regular day across three frequency bands managed by the eNB and different time periods (morning, afternoon, evening, night).}}
	\label{fig:PRBscatter}
\end{figure*}

\setcounter{figure}{3}
\begin{figure}[t!]
\centering
\includegraphics[width=\columnwidth, clip, trim = 0cm 0cm 0cm 0cm]{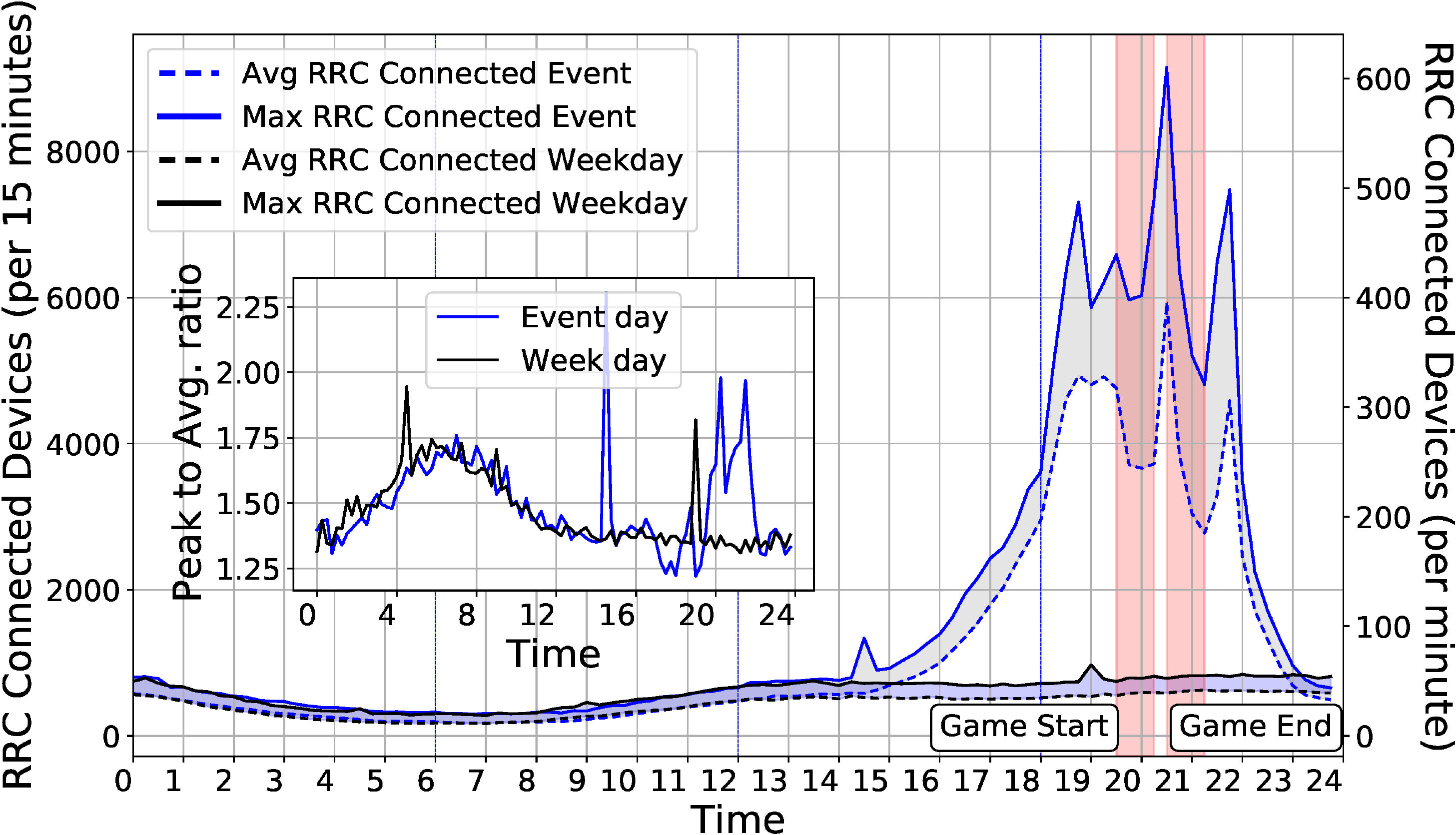}
\caption{Evolution of users activity. Peak and average number of active users ($\texttt{RRC}_\texttt{CONNECTED}$) (outer plot) and peak-to-average ratio (inner plot) during the day of the event and during a regular day.}
\label{fig:RRC}
\end{figure}

\subsection{Traffic Volume Patterns}
Fig.~\ref{fig:traffic_volumes} shows the aggregated traffic volume generated within a $24$-hour time period of a regular weekday (dashed line) and also during the day of the event (straight line), discerning between uplink (UL) and downlink (DL) traffic with red and blue lines, respectively. The event is scheduled to start at 19.30h and finish at 21.15h. We first observe that both UL and DL patterns are remarkably similar between a regular day and the day of the event up until 3 hours before the beginning of the event and the anomaly data usage persists until around 2 hours after the end of the event. This clearly reflects common human behavior as attendees usually gather in the surroundings of the stadium for social interactions before and after the event. This is evidenced by two volume peaks at around 18h and at around 21.30h. The former gradually vanishes until the event begins, coinciding with crowds moving towards the allotted seats in the stadium, whereas the latter peaks practically immediately after the end of the event and gradually vanishing as crowds move out of the stadium. Interestingly, there is a third peak, concomitant with the break of the event, with two drops in volume during the two halves of the event, which is expected as users are focusing on the game itself. We would like to remark that this pattern repeats across all events of the season with a gain factor dependent on the number of attendees and represents a signature of football games. Different events will leave different footprints and motivate the use of model-free approaches to forecast capacity requirements.

Fig.~\ref{fig:scatter_volumes} compares the same distributions of traffic volume across different time periods, namely, morning (6-12h), afternoon (12-18h), evening (18-24h) and night (24-6h), as a function of the number of active users ($\texttt{RRC}_\texttt{CONNECTED}$) for both downlink and uplink in a regular weekday (first and second plots) and for the day of the event (third and fourth plots). As expected, the volume distribution during the day of the event is skewed with a long tail concentrating large number of users during the afternoon and the evening, precisely the new connections from the event attendees. This allows us to characterize the background traffic volume, generated by less than $250$ active connections spawning less than $50\%$ of the total system capacity following a strongly linear relationship. Such linearity is broken during the event for downlink traffic. The average data volume by active users within this regime (above $250$ users) is lower than in normal conditions, which suggests poor quality of service experienced by the end users in the stadium.

\subsection{Temporal Distribution of Mobile Users}
\label{subsec:TemporalDistribution}

\setcounter{figure}{5}
\begin{figure}[!t]
\centering
\includegraphics[width=\columnwidth, clip, trim = 0cm 0cm 0cm 0cm]{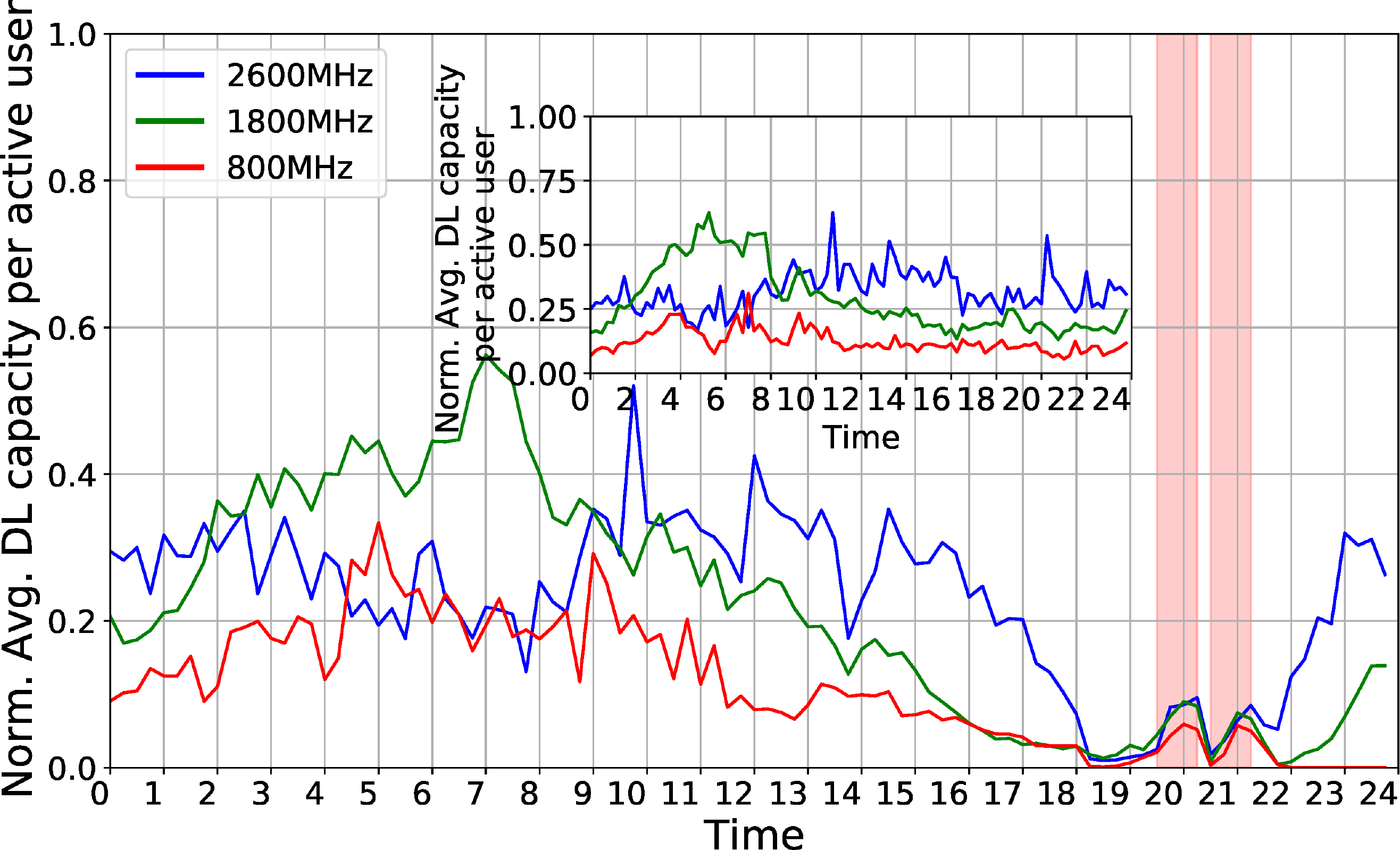} 
\caption{Evolution of the average downlink (DL) capacity per user (our QoS metric) during the day of the event (outer plot) and a regular day (inner plot). Highlight areas indicate the two periods of time when the sport event is occurring.}
\label{fig:Capacity}
\vspace{-2mm}
\end{figure}

Fig.~\ref{fig:RRC} compares the activity of active users ($\texttt{RRC}_\texttt{CONNECTED}$) in the neighbourhood of the stadium during the day of the event (blue lines) with that of a regular weekday (dark lines), aggregated over all the eNBs in the area. The figure shows the average number of users within each $15$-minute time window (dashed lines) and the maximum number of active users within the same window (straight lines). As expected, the number of active connection peaks are highly correlated to the traffic volume time evolution shown in Fig.~\ref{fig:traffic_volumes}, with the number of concurrent connections ranging up to $600~\frac{\textup{active UEs}}{\textup{minute}}$, over $10$ times higher than those during a regular day. The areas highlighted in light blue between straight and dashed lines provide an indication of the variability of the number of active UEs, i.e., the peak-to-average active connections, and evidence a substantial increase in volatility (around $2$ times higher peak-to-average connections) during the course of the event. To visualize this, we create an inner plot depicting the peak-to-average ratio in both scenarios (event and regular day). Higher volatility in user activity is observed during the night and early in the morning, which is explained by the low number of net active connections (small variations yield higher peak-to-average ratios). Interestingly, however, the average-to-peak ratio increases during the event, which implies that the operator must deal with substantial control plane volatility in addition to data volume surges. 
It is clear that as soon as the base station reaches the saturation point, the perceived QoE decreases due to limited scheduling of resources. We further investigate this point in the following subsections.

\begin{figure*}[t!]
\centering
  \begin{subfigure}[b]{.50\textwidth}
    \includegraphics[width=\textwidth, height=10cm, keepaspectratio ]{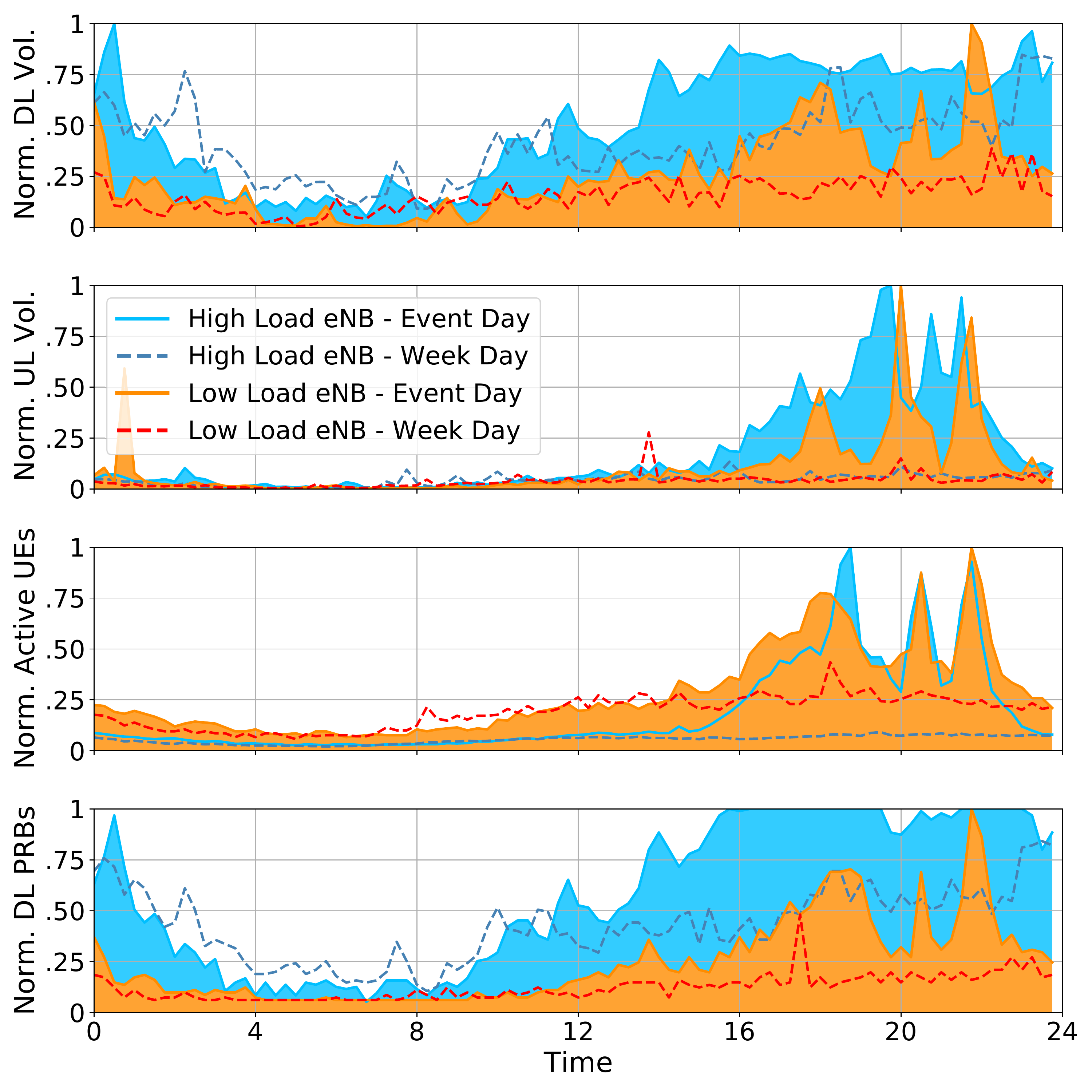}
    \caption{Overview of different metrics collected by two eNBs in the stadium area during an event day and weekday (Normalized values).}
    \label{fig:TimeSeries}
  \end{subfigure}\qquad
  \begin{subfigure}[b]{.45\textwidth}
    \includegraphics[width=\textwidth, clip, trim = 0cm 0cm 0cm 0cm,height=5.5cm, keepaspectratio]
    {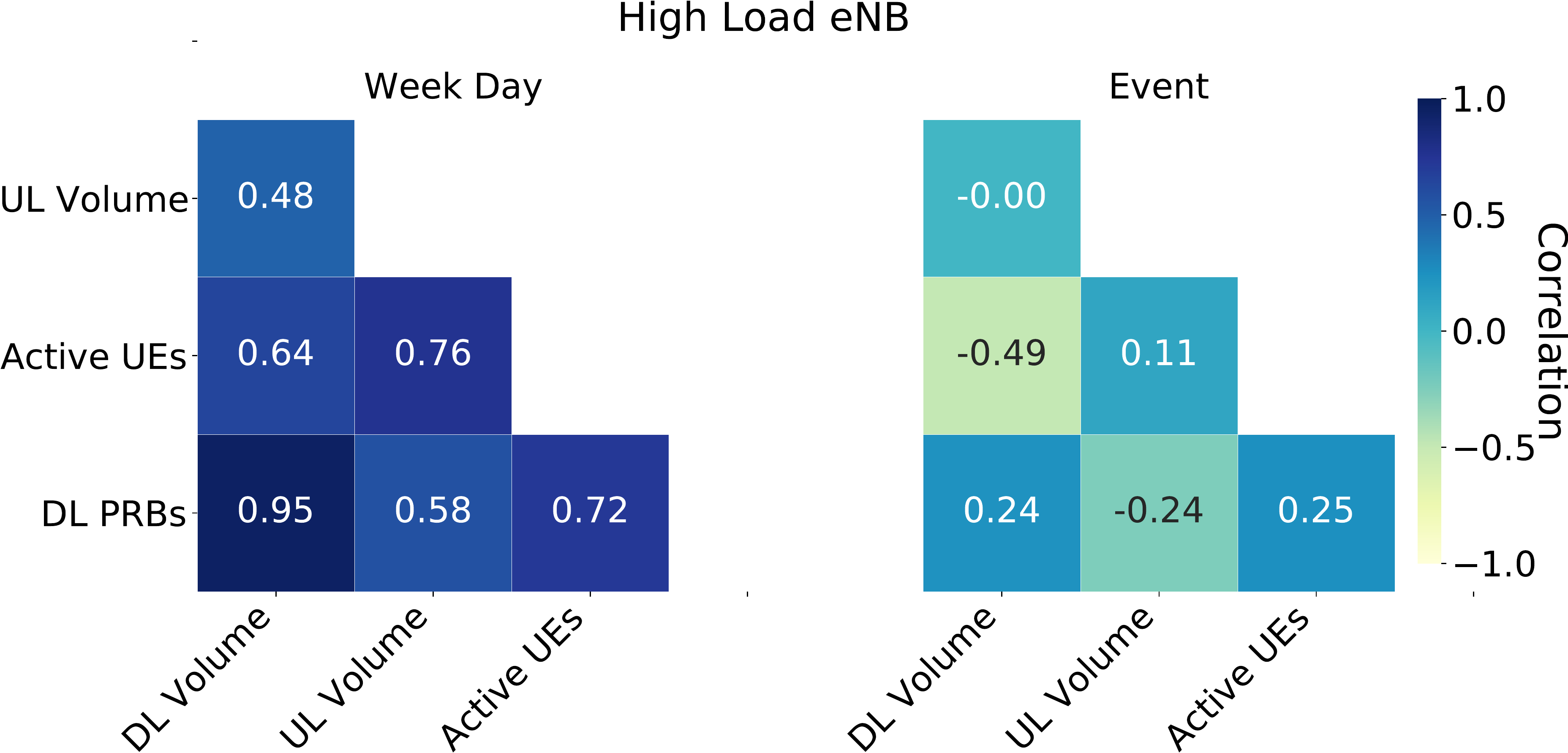}
    \caption{Correlation matrix for different traces collected by a highly loaded eNB during a weekday (left) and the event (right).}
    \label{fig:Correlation_NormalDays}
  \vspace{1ex}
    \includegraphics[width=\textwidth, clip, trim = 0cm 0cm 0cm 0cm,height=5.5cm, keepaspectratio]
    {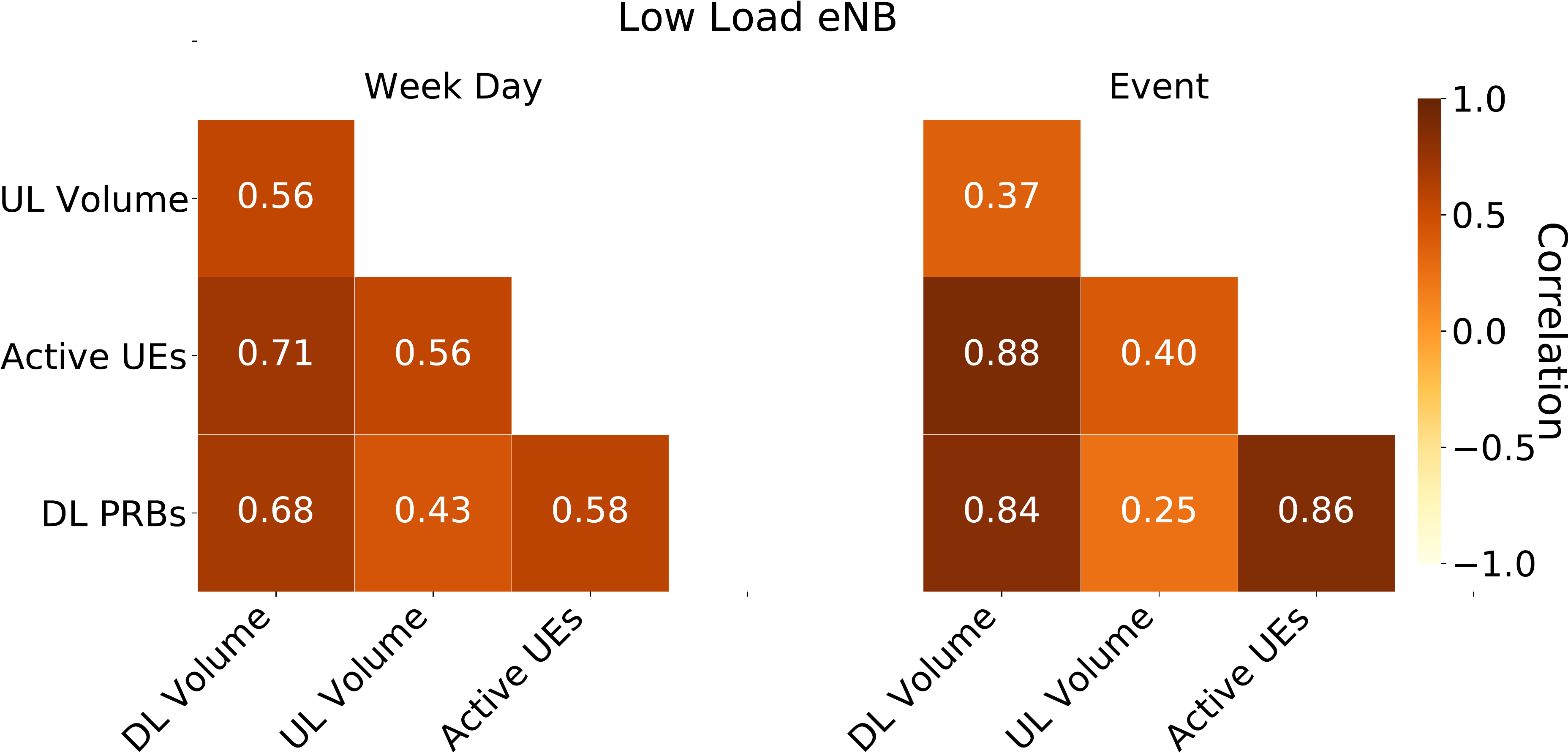}
    \caption{Correlation matrix for different traces collected by a lightly loaded eNB during a weekday (left) and the event (right).}
    \label{fig:Correlation_Event}
  \end{subfigure}%
\caption{Temporal correlation of features in different working conditions.}
\label{fig:correlation}
\vspace{-3mm}
\end{figure*}

\subsection{Service Degradation}
\label{subsec:ServiceDegradation}

As mentioned above, we conjecture that the skewness of the volume distribution shown in Fig.~\ref{fig:scatter_volumes} is due to poor quality of service. To confirm this hypothesis, we take a closer look to the physical spectrum usage, namely the downlink physical resource block (PRB) utilization measured at the eNBs. In this way, Fig.~\ref{fig:PRBscatter} compares the relative PRB utilization as a function of the number of active users during the event (first and third plot) and that of a regular day (second and fourth plot) for the different frequency bands used by the sectors of the eNBs (first two plots) and for different periods of time during the day (second two plots). We note that the $1800$-MHz band saturates when the number of users grows over $250$, which unavoidably translates into QoS degradation for the end users. The time analysis confirms that this event occurs during the time period of the football match.

This is further confirmed in Fig.~\ref{fig:Capacity}, which compares the average throughput capacity per user over time during the day of the event (outer plot) against a regular weekday (inner plot), considering each of the three bands selected for our study. As expected, individual QoS substantially deteriorates during the sport event, particularly during a 2.5-hour time window before the beginning of the event, during the break of the event and during a 1.5-hour time window after the end of the event. With respect to standard working conditions, the average throughput capacity per user decreases up to $50$ times.

\subsection{Inter-Feature Time Correlation}
\label{subsec:TimeCorrelation}
We now study how different features relate to each other and across different neighbouring eNBs from a statistical perspective. To ease the presentation, we focus on two eNBs: an eNB that carries significant traffic volumes during both week and event days, labelled as "High Load eNB" and represented with a blue color palette in Fig.~\ref{fig:correlation}; and an eNB that serves less significant amount of traffic, labelled as "Low Load eNB" and represented with a red color palette. 
Specifically, Fig.~\ref{fig:TimeSeries} shows the temporal dynamics of different features collected by the two eNBs during a regular weekday (dashed lines) and during a match game (shaded area), while Figs.~\ref{fig:Correlation_NormalDays} and \ref{fig:Correlation_Event} depicts their correlation matrix.
\change{From Fig.~\ref{fig:TimeSeries}, it clearly appears the performance gap with standard operating conditions. Specially for the UL case, during the event both the "High Load" and "Low Load" eNBs present traffic peaks $10$ times higher than the traffic volumes achieved in normal conditions. Such UL activity peak can be justified by a sudden increase of social media activity and background cloud synchronization processes~\cite{Superbowl2013}. }

Let us first focus on the left matrices of Figs.~\ref{fig:Correlation_NormalDays} and \ref{fig:Correlation_Event}. As expected, since both the eNBs are deployed in the same area and the human activity corresponds to that of a regular weekday, the correlation value (Pearson's r coefficient) between each pair of features is positive and similar in both eNBs. This is true for every couple of features except the pair "DL PRBs-DL Volume", which appears strongly correlated for the highly loaded eNB and only moderately correlated for the lightly loaded eNB.
It should be noticed that the latter exhibits a rather flat behavior during normal working conditions.
More interesting are the correlation matrices we observe during the match, depicted by the two right-most plots of Figs.~\ref{fig:Correlation_NormalDays} and \ref{fig:Correlation_Event}. First, the lightly loaded eNB shows higher correlation values with respect to normal working conditions due to a more dynamic pattern enforced by the attendees in the area.
Second and more surprisingly, we observe that most of the correlations available during a weekday in the highly loaded eNB vanish during the match, eventually providing negative value for the couple "Active UEs-DL Volume". The latter represents a counter-intuitive behaviour explainable by the fact that the considered eNB saturates its physical resources (PRBs) during most part of the game distorting the underlying relationship between the features. 
Together, these observations provide valuable insights not only on the fact that in normal working conditions features such as Active UEs and DL Volume are correlated (which is somewhat expected), but also that during mass events distortions may occur. 
This motivates us to design a mechanism that exploits not only the information  contained  within  each  time series feature, but also the time-domain correlation that heterogeneous sequences might show among  themselves. Moreover, as each eNB exhibits different behaviour, we advocate to model each cell individually taking into account both regular and crowded situations.

%

\section{Model Design}
\label{sec:model}

\change{In our data analysis, we showed that current network deployments may be insufficient during football events as the unexpected-high density of users can quickly saturate the available network resources. However, providing additional spectrum via increasing the density of radio access points is overly expensive, even with future network slicing mechanisms.} Hence, proper mechanisms to accurately account for the required additional resources so that certain QoS targets are satisfied must be in place. 
Hereafter, we model the problem of network configuration during \change{football} events and suggest a new network setting to be dynamically applied by the operator.

\subsection{Notation}

We use conventional notation.
We let $\mathbb{R}$ and $\mathbb{Z}$ denote the set of real and integer numbers. 
We use $\mathbb{R}_+$, $\mathbb{R}^n$, and $\mathbb{R}^{m\times n}$ to represent the sets of non-negative real numbers, $n$-dimensional real vectors, and  $m\times n$ real matrices, respectively. 
Vectors are in column form and written in bold font. 
Matrices are in upper-case font. 
Subscripts represent an element in a vector and superscripts elements in a sequence. For instance, $\langle \vvi x{t} \rangle$ is a sequence of vectors with $\vvi x{t} =   ( x^{(t)}_1, \dots,  x^{(t)}_n   )^T$ being a vector from $\mathbb{R}^n$. In turn, $x^{(t)}_i$ is the $i$'th component of the $t$'th vector in the sequence, where superscript $T$ represents the transpose operator.
$\| \vv x \|_2$ represents the 2-norm or Euclidean norm of $\vv x$ and $\norm{\vv x}_\infty$ its maximum norm ($\max_i {|x_i|}$).
Finally, $\left [\cdot\right ]^+$ denotes the projection of a vector onto the non-negative orthant, i.e., $\left [\vv x \right ]^+ = \left ( \max\{0,x_1\},\dots,\, \max\{0,x_n\}\right ), x\in\mathbb{R}^n$.

\subsection{Problem definition}
Let us denote the set of base stations in the considered area as $b\in\mathcal{B}$, where $B=|\mathcal{B}|$ is the total amount of base station sectors. Without loss of generality, we consider each sector as an independent base station as each sector shows heterogeneous behaviour and each of them cover a different subset of the users in the area. We assume that time is split into different epochs $e\in E$ with duration $T$. Let $r_u^e$ denote the amount of traffic served for user $u\in\mathcal{U}_b^e$ during epoch $e$ (bits), where $U_b^e=|\mathcal{U_b}^e|$ is the average number of users under the coverage area of base station $b$ connected concurrently within epoch $e$. Each base station $b$ is provided with a spectrum capacity $c_b^e$ (number of PRBs), which can be dynamically changed (every epoch $e$) based on the carrier aggregation policies of the telco operator. Note that in case of seriously-congested events, the operator may even decide to augment the overall spectrum capacity by placing portable base stations (e.g., Cells on Wheels or Cells on Light Trucks) or leasing or sharing a portion of the available bandwidth from other operators via, e.g., network slicing~\cite{Sciancalepore_TNET19}.
We now characterize the satisfaction of individual user connections by formally introducing the QoS metric used in Fig.~\ref{fig:Capacity} as the following expression:
\begin{align}
    &q_b^e = \frac{\sum_{u\in\mathcal{U}_b^e}r_u^e}{ \sum_{u\in\mathcal{U}_b^e}{\tau_u^e}}, &\forall b\in\mathcal{B}.
\end{align}
where $r_u^e$ represents the volume of traffic per user under the base station coverage served within the epoch $e$, and $\tau_u^e$ the effective downlink time per user, which accounts for the time when the first part of the PDCP SDU of the downlink buffer was transmitted until the buffer is emptied. 
It should be noticed that in presence of congestion and fair scheduling of resources, the average active time per user increases as a result of resource contention, thus leading to lower QoS. However, $\tau_u^e$ depends on the spectrum capacity covering the event \emph{and} on the contextual data around the event (type of event, expected number of attendees) and the model capturing the behavior of $\tau_u^e$ is not known \emph{a priori} and so is $q_b^e$'s. 

{\it Objective.} The target is to find the best PRB allocation on different deployed base stations such that the overall cost of such additional spectrum is minimized such that the aggregated QoS measured from all considered base stations is above a certain pre-defined threshold $\Gamma$. Of course, our focus is on the amount of spectrum required during mass events, which will typically exceed the capacity of the system and depends on the context of the event itself.

{\it Constraints.} We consider two sets of constraints. First, we need to impose that traffic queues are stable, that is, they do not grow infinitely over time, which would lead to dramatic drops in latency performance. We do not impose an instantaneous serving rate greater than the arrival rate; instead we consider that the long-term base station serving rate must be greater or equal to the long-term data arrival rate. Formally, this can be expressed by
\begin{align}
    &\sum\limits_{e\in\mathcal{E}}\sum\limits_{u\in\mathcal{U}_b^e} \left( r_u^e - \zeta_u^e(c_b^e)\right) \leq 0,\quad\forall b\in\mathcal{B}.
\end{align}
where $\zeta_u^e$ is the average throughput over epoch $e$ for user $u$. Second, we shall impose certain threshold on individual QoS performance, which in turn can be succinctly described by
\begin{align}
\sum\limits_{e\in\mathcal{E}}\left(\Gamma_b^e-q_b^e(c_b^e)\right) \leq 0,\quad\forall b\in\mathcal{B}
\end{align}
where $-q_b^e(\cdot)$ is the QoS of base station $b$ during epoch $e$. While both functions $q_b^e(\cdot)$ and $\zeta_u^e(\cdot)$ are not known, we can get estimates at the end of each epoch $e$.

{\it Convex functions and decision variables.} Telco operators may apply different base station settings, i.e., the number of available spectrum resources (PRBs), and decide to increase the total number of available spectrum in order to tune users' performance. We assume that the functions capturing throughput and QoS performance are both convex with respect to the amount of spectrum or number of selected PRBs (the larger the spectrum, the higher the user throughput, the better the QoS). Note that a trivial solution to satisfy all user requirements, i.e., to keep the QoS at reasonable levels, is to provide an infinitely large amount of spectrum. However, deploying additional spectrum (e.g. via CoWs or network slicing) incurs in a cost $\delta^e$ for the operator that we shall minimize.

\subsection{Bandit convex optimization}
With the above information, we can mathematically formulate the following optimization problem
\begin{align}
		\min\limits_{\vv{c}\in\mathbb{N}^{B\times E}}& \sum\limits_{e\in\mathcal{E}}\sum\limits_{b\in\mathcal{B}}\delta^e(c_b^e)\\
		\st{&\sum\limits_{e\in\mathcal{E}}\sum\limits_{u\in\mathcal{U}_b^e} \bigl( r_u^e - \zeta_u^e(c_b^e)\bigr) \leq 0},\quad\forall b\in\mathcal{B};\nonumber\\
        &\sum\limits_{e\in\mathcal{E}}\bigl(\Gamma_b^e-q_b^e(c_b^e)\bigr) \leq 0,\quad\forall b\in\mathcal{B};\nonumber\\
        &c_b^e\in\mathbb{R}^{+}, \quad\forall b\in\mathcal{B}, e\in\mathcal{E};\nonumber
\end{align}
where $\zeta_u^e$, $q^e_b$ and $\delta^e$ depend on the amount of spectrum (PRBs) $c_b^e$ allotted to base station $b$ during epoch $e$.  We assume the following cost function
\begin{align}
    \delta^e(c_b^e)= \begin{dcases}
    0, \qquad \text{if } c_b^e \leq 1,\\
    \alpha\, (c_b^e)^\beta, \quad\text{if } c_b^e > 1;
\end{dcases}
\end{align}
so $\delta(c_b^e)$ captures the fact that base stations might be assigned with PRBs already available within the bandwidth provisioned to the ($c_b^e\leq 1$) and so it does not incur in extra cost, or additional spectrum is required ($c_b^e>1$) with a cost parametrized by $\alpha$ and $\beta$. 

Recalling the online convex optimization theory~\cite{Chen2017TSP} and considering an invariant feasible set $\mathcal{C}=(c_b^e)$, we can rewrite the above optimization problem with its online Lagrangian version as follows
\begin{equation}
    \mathcal{L}_e(\vv{c}^e,\vv{\lambda}):=\delta^e(\vv{x})+\vv{\lambda}^T\vv{g}^e(\vv{x}),
\end{equation}
where $\vv{c}^e=(c_b^e)$ is the set of selectable PBRs during epoch $e$ and $\vv{g}^e(\vv{c}^e) = \sum\limits_{e\in\mathcal{E}}\left(\vv{\Gamma}^e-\vv{q}^e(\vv{c}^e) + \vv{r}^e - \vv{\zeta}^e(\vv{c}^e)\right)$.
We can then write the recursive $\vv{c}^{e+1}$ given the prime iteration $\vv{c}^e$ as the following
\begin{equation}
    \vv{c}^{e+1} = \arg\min\limits_{\vv{x}\in\mathcal{C}} \nabla_{\vv{x}}^T\mathcal{L}^e(\vv{c}^e,\vv{\lambda}^e)(\vv{x}-\vv{c}^e)+\frac{1}{2\sigma}||\vv{x}-\vv{c}^e||^2,
\end{equation}
where $\sigma$ is a predefined constant and $\nabla_{\vv{x}}^T\mathcal{L}^e(\vv{c}^e,\vv{\lambda}^e)=\nabla\delta^e(\vv{c}^e+\nabla\vv{g}^e(\vv{c}^e)\vv{\lambda}^T$. This admits the following solution:
\begin{equation}
\label{eq:primate}
    \vv{c}^{e+1} = \mathcal{P}_{\mathcal{C}}\left(\vv{c}^e -\sigma\nabla_{\vv{x}}\mathcal{L}^e(\vv{c}^e,\vv{\lambda}^e)\right),
\end{equation}
where the projector operator is $\mathcal{P}_{\mathcal{C}}(\vv{y}) = \arg\min\limits_{\vv{c}\in\mathcal{C}}||\vv{c}-\vv{y}||^2$, and the dual update as the following
\begin{equation}
\label{eq:dual_eq}
    \vv{\lambda}^{e+1}\!\! = \!\!\left[\vv{\lambda}^e \! +\! \mu\left(\vv{g}^e(\vv{c}^e)+\nabla^T\vv{g}^e(\vv{c})^e)(\vv{c}^{(e+1)}\!\!-\vv{c}^e)\right)\right]^+,
\end{equation}
where $\mu>0\in\mathbb{R}^+$ is a step size.

Given that functions $\zeta(\cdot)$ and $q(\cdot)$ are not known, we need to construct a stochastic gradient estimate of unknown functions using the limited value information~\cite{SODA2005_flaxman}.  In particular, we can evaluate the function at a perturbated point $\vv{x} + \epsilon\vv{u}$ yielding that $\nabla f(\vv{x})\approx E[\hat{\nabla}^1f(\vv{x})]$, where $\hat{\nabla}^1f(\vv{x})= \frac{d}{\epsilon}f(\vv{x}+\epsilon \vv{u})\vv{u}$ is the one-point gradient~\cite{duchi2015}. Therefore, we need to recall the bandit online optimization theory to rewrite Eq.~\eqref{eq:primate} as follows
\begin{equation}
    \hat{\vv{c}}^{e+1} = \mathcal{P}_{(1-\epsilon)}\mathcal{C}\left( \hat{\vv{c}}^e - \sigma\hat{\nabla}^1_{\vv{c}}\mathcal{L}^e(\vv{c}^e,\vv{\lambda}^e) \right),
\end{equation}
where the projector operation is performed within a subset of $\mathcal{C}$ to ensure feasibility of the perturbed $\hat{\vv{c}}^e$. We can then write the dual update operation (Eq.~\eqref{eq:dual_eq}) as the following 
\begin{equation}
    \vv{\lambda}^{e+1}\!\! = \!\!\left[\vv{\lambda}^e \! +\! \mu\left(\vv{g}^e(\hat{\vv{c}}^e)+\nabla^T\vv{g}^e(\hat{\vv{c}}^e))(\hat{\vv{c}}^{(e+1)}\!\!-\hat{\vv{c}}^e)\right)\right]^+,
\end{equation}
where $\hat{\vv{c}}^e$ is the learning iterate.

\change{The above considerations suggest that the model may fail while trying to approximate functions $\zeta(\cdot)$ and $q(\cdot)$. 
In addition, we need the exact characterization of the mobility patterns influencing the 
number of connected user under each base station $\mathcal{U}_b^e$, for each time epoch $e$ and base station $b$, as well as the expected traffic demand $r_u^e$. 
Unfortunately, all these variables present a strong relationship with respect to multiple real-world variables, making our objective hardly achievable within a reasonable time window. Therefore, in the next section, we propose a machine-learning-based approach to automatically learn and approximate these functions, and provide forecasts on the number of active users as well as their traffic request patterns based on a data-driven approach.
}

\section{\name{}: Design and Performance}\label{sec:ML}
Our solution design, \name{}, requires two different ML-based models to estimate the number of active users and their traffic patterns. On the one hand, an LSTM-based approach allows to predict repetitive time patterns. On the other hand, a Deep-Learning model allows to approximate complex functions, e.g., those related to human mobility and cellular traffic loads. Thus, we first focus on inferring the number of active users in the stadium area, and then we exploit this information to estimate the additional resource utilization necessary to guarantee normal-condition QoE. Finally, we show the performance of \name{} applied to real traffic data.

\begin{figure*}[!t]
\centering
  \begin{subfigure}[b]{.49\textwidth}
    \centering
    \includegraphics[width=\columnwidth, clip, trim = 0cm 0cm 0cm 0cm]{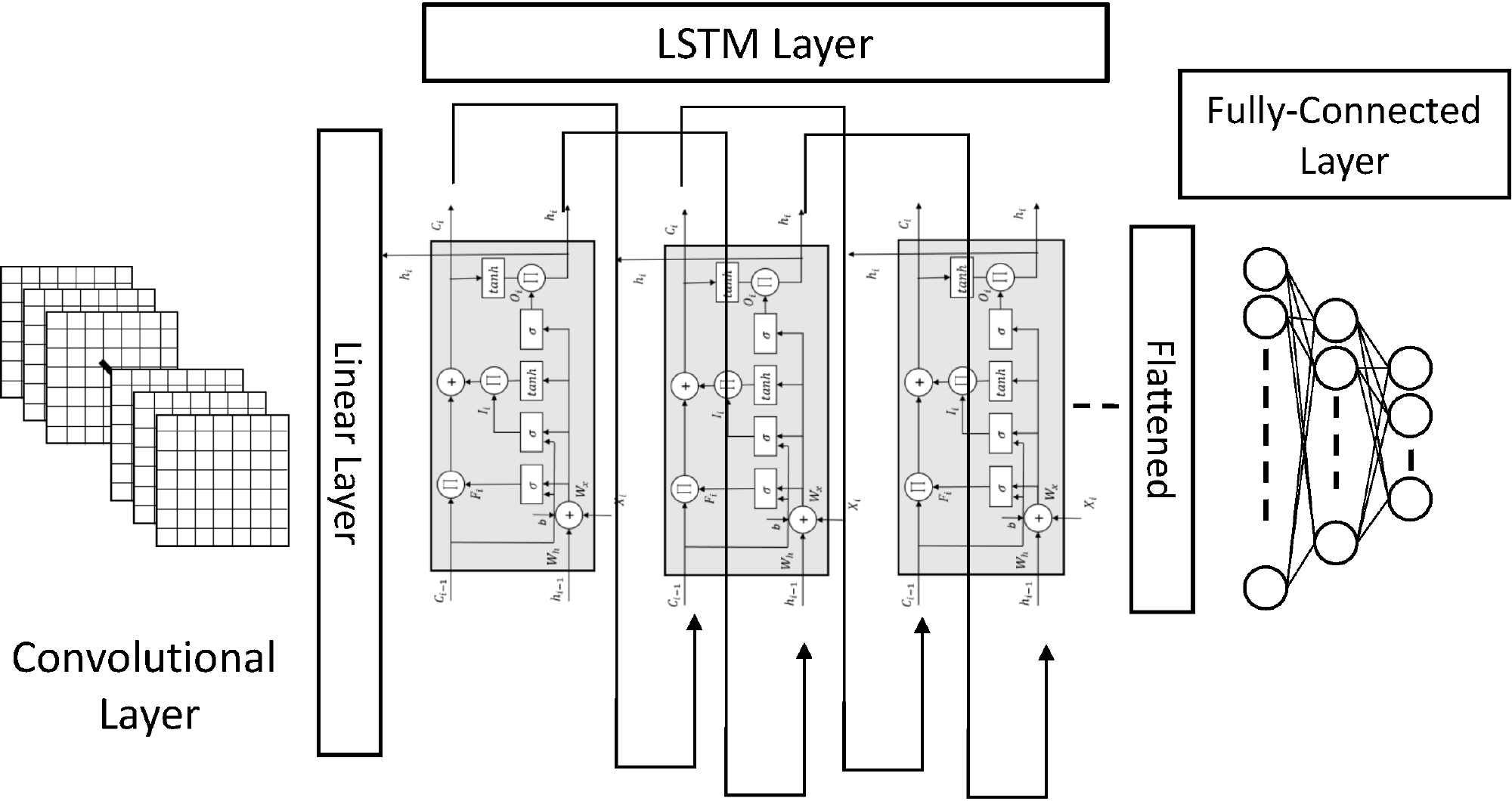}
    \caption{\change{CNN-LSTM architecture.}}
    \label{fig:cnn-lstm}
  \end{subfigure}%
  \begin{subfigure}[b]{.47\textwidth}
  \centering
    \includegraphics[width=\columnwidth,trim = 0cm 0cm 0cm 0cm,clip, height=5cm, keepaspectratio]{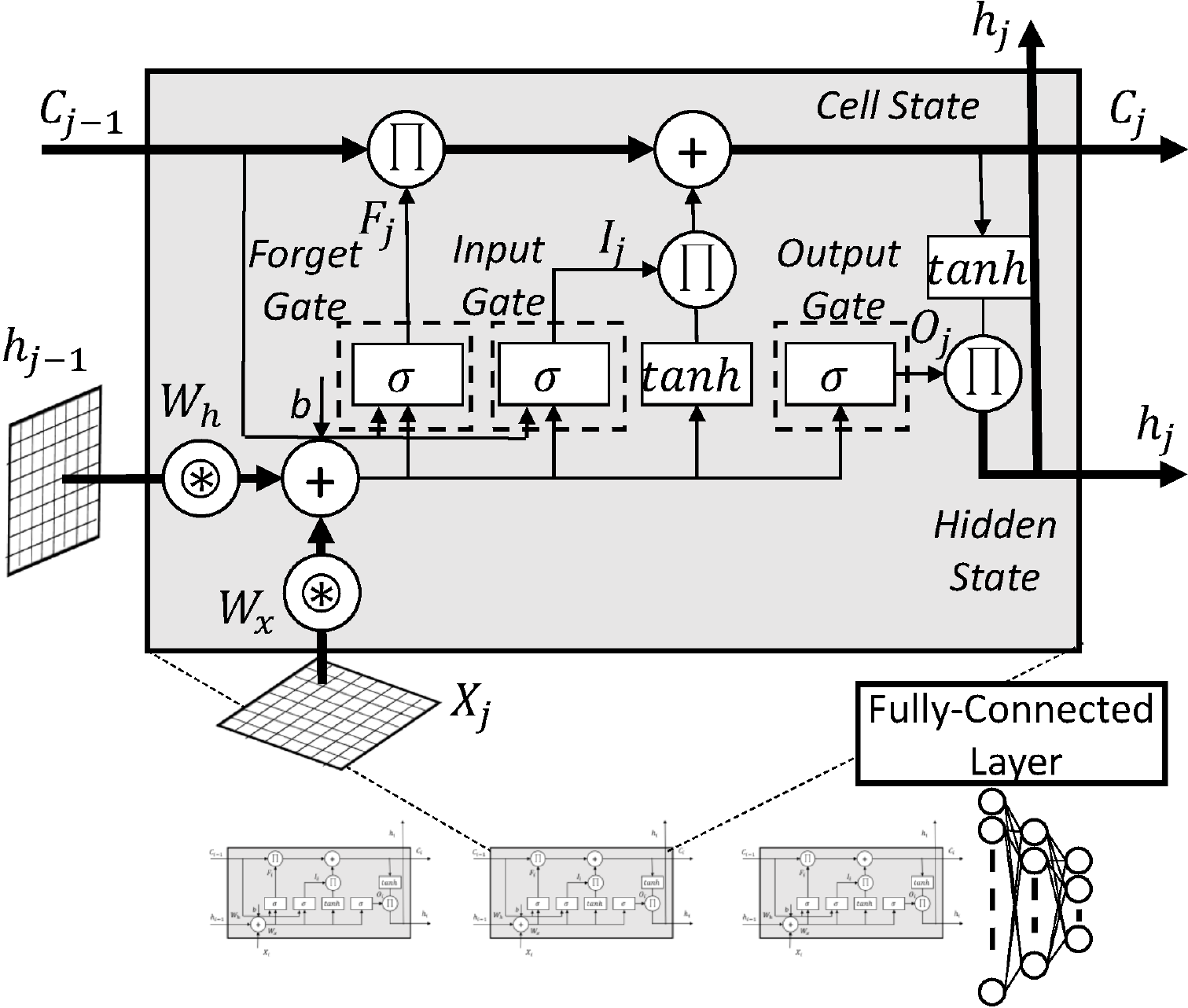}
    \caption{\change{ConvLSTM architecture.}}
    \label{fig:Conv_lstm}
  \end{subfigure}%
\caption{Comparison of different LSTM-based architectures.}
\vspace{-5mm}
\label{fig:lstm_comparison}
\end{figure*}

\subsection{Active users}
\label{subsec:Active_users}

Spatio-temporal characteristics of mobile traces unveil patterns that might be exploited to perform a forecasting process. In this context, Long Short-Term Memory Recurrent Neural Networks (LSTM-RNNs)~\cite{LSTM} have attracted attention given their capacity to capture repetitive schemes within unstructured time series.
An LSTM network is composed by a chain of units, which sequentially apply a linear combination of operations on the input data. Such a structure is fundamental to provide the network with the capability to~\emph{remember} useful information gathered from the past data samples and learn long-term trends in the input sequence making LSTM well-suited to handle mobile data traffic with significant spatio-temporal correlations~\cite{gramaglia_data_orange}.
LSTMs can be represented as a chain of $J$ units, where each LSTM unit $j \in \mathcal{J}$ consists of $4$ main elements, a memory cell and three gates, namely the input $I_j$, the output $O_j$ and the forget gate $F_j$. The input and output gates are responsible to insert and read back the data into/from the memory cell itself, while the forget gate determines how much details should be kept or removed from the unit at each iterative step. The information stored within each unit $j \in \mathcal{J}$ is commonly referred as ``state'' of the unit, or $C_j$. To build more complex models, LSTM units are usually stacked and concatenated into consecutive layers, where the output of one layer feeds the input of the subsequent one, and the hidden state $h_j$ of each unit holds the information about previously observed data. During the training phase, the input and the previous hidden state are combined to form a vector, which holds current and past information. Then, the vector passes through several activation functions, namely sigmoid and hyperbolic tangent functions depicted as $\sigma$ and $tanh$ in Fig.~\ref{fig:Conv_lstm}, which help regularizing the outcome of the linear combinations, finally adapting weight $W$ and bias $b$ values of each gate. 
Although the LSTM model has proven powerful in handling temporal correlation, existing work in the literature suggests the use of LSTMs together with Convolutional Neural Network (CNNs) in order to increase the accuracy of the learning models~\cite{Zhang2018}.
The key-idea is to exploit not only the information contained within each time series, but also the time-domain correlation heterogeneous sequences might show among themselves, as discussed in Section~\ref{subsec:TimeCorrelation}.
For this purpose, we consider two well-known architectures, namely CNN-LSTM~\cite{Learning_Traffic_As_Images} and ConvLSTM~\cite{ConvLSTM}, and compare their capabilities in predicting the number of $\texttt{RRC}_\texttt{CONNECTED}$ devices against the legacy LSTM model within a week time span. 

As depicted in Fig.~\ref{fig:lstm_comparison}, the main difference about the two models is that ConvLSTM embeds the convolution steps into the LSTM unit, while CNN-LSTM models imply the concatenation of the two types of networks sequentially\footnote{We refer the reader to state-of-the-art literature such as~\cite{Learning_Traffic_As_Images} and~\cite{ConvLSTM} for further implementation details.}.
Both approaches therefore require the input data (and the output of the internal transfer functions) to be shaped as tensors to take advantage of the 2D representations of the network activities and include spatio-temporal information to learn valuable patterns within the dataset.
\change{The overall dataset has been split according to a $60/40$ ratio for the purposes of training and validation, respectively.} We train the models exploiting a moving window of $6$ days as input data and inferring on the consecutive one, over a time span of several weeks.
\change{ Within our experimental setup, we considered several hyperparameter settings. In particular, we account for a variable number of LSTM cells $J =\{ 200,300,400\}$ and convolutional filters $\Phi =\{ 64,128,256\}$ for each model architecture.
The legacy LSTM model is composed of two consecutive layers of $J$ units and a final fully-connected layer that provides the resulting output.
The CNN-LSTM model accounts for two convolutional layers characterized by $\Phi$ filters each and a kernel size of $3$ to learn spatial features, which are followed by an LSTM layer of $J$ units and a final fully-connected layer to learn the correlation in time.
With respect to legacy LSTM models, the ConvLSTM architecture neglects the internal dense connection among gates in favour of a convolution operation, thereby reducing the number of model parameters and decreasing the chances of over-fitting.
We adopt the mean-squared error (MSE) metric to train the models, and choose the Adam optimizer to optimize the loss function~\cite{vani2019}. 
For the sake of comparison, we additionally include a baseline autoregressive integrated moving average (ARIMA) model characterized by parameters $(p,d,q)$, where $p$ is the order of the autoregressive term, $d$ is the number of differencing required to make the time series stationary, and $q$ is the moving average order~\cite{ARIMA}.
This model requires the input time series to be stationary. Therefore, we applied simple differencing techniques to satisfy this requirement. Additionally, we implemented a grid search algorithm to optimize the ARIMA parameters choice, leading to the following settings $(p,d,q)=(5,1,0)$. We resume in Table~\ref{tab:Errors} the resulting MSE and maximum absolute error measured over the forecasted sequences, averaged over $4$ weeks of testing data. As expected, due to the non-stationarity of traffic traces (demonstrated by sudden peaks during the most busy time periods) the ARIMA predictor provides weak performances when dealing with the event time period, resulting in a maximum absolute error of $0.505$. From our findings, we can conclude that the CNN-LSTM model with $J$=$200$ and $\Phi$=$64$ provides best performances in this scenario. Hence hereafter, we will adopt this model as our active user predictor.
}





\begin{table}[!b]
\footnotesize
\centering
\caption{\small \change{Error characterization over different settings and models.}}
\label{tab:Errors}
\resizebox{\columnwidth}{!}{\begin{tabular}{|c|c|c|c|c|c|}
\hline
                            & \textbf{ J / $\Phi$ } & \textbf{LSTM} & \textbf{CNN-LSTM} & \textbf{ConvLSTM} & \textbf{ARIMA} \\ \hline
\multirow{3}{*}{MSE}      & $64$/$200$    & $0.0039 $  & $0.0032 $  & $0.0036 $  &    \multirow{3}{*}{$0.055$}\\ \cline{2-5} 
\multicolumn{1}{|c|}{}      & $128$/$300$    & $0.0041 $  & $0.0035 $  & $0.0040 $ &  \multicolumn{1}{c|}{} \\ \cline{2-5} 
\multicolumn{1}{|c|}{}      & $256$/$400$    & $0.0045 $  & $0.0039 $  & $0.0037 $  & \multicolumn{1}{c|}{} \\ \hline \hline

\multirow{3}{*}{ \vtop{\hbox{\strut MAX ABS}\hbox{\strut Error}} }    & $64$/$200$     & $0.321 $ & $0.339 $  &  $0.322 $ &  \multirow{3}{*}{$0.505$}\\ \cline{2-5} 
                            & $128$/$300$ & $0.296 $  & $0.322 $  & $0.318 $ & \multicolumn{1}{c|}{}\\ \cline{2-5} 
                            & $256$/$400$     &  $0.321 $  &  $0.328 $   & $0.321 $  & \multicolumn{1}{c|}{} \\ \hline
\end{tabular}}
\end{table}

\begin{figure}[!t]
\centering
\includegraphics[width=\columnwidth, clip, trim = 0cm 0cm 0cm 0cm]{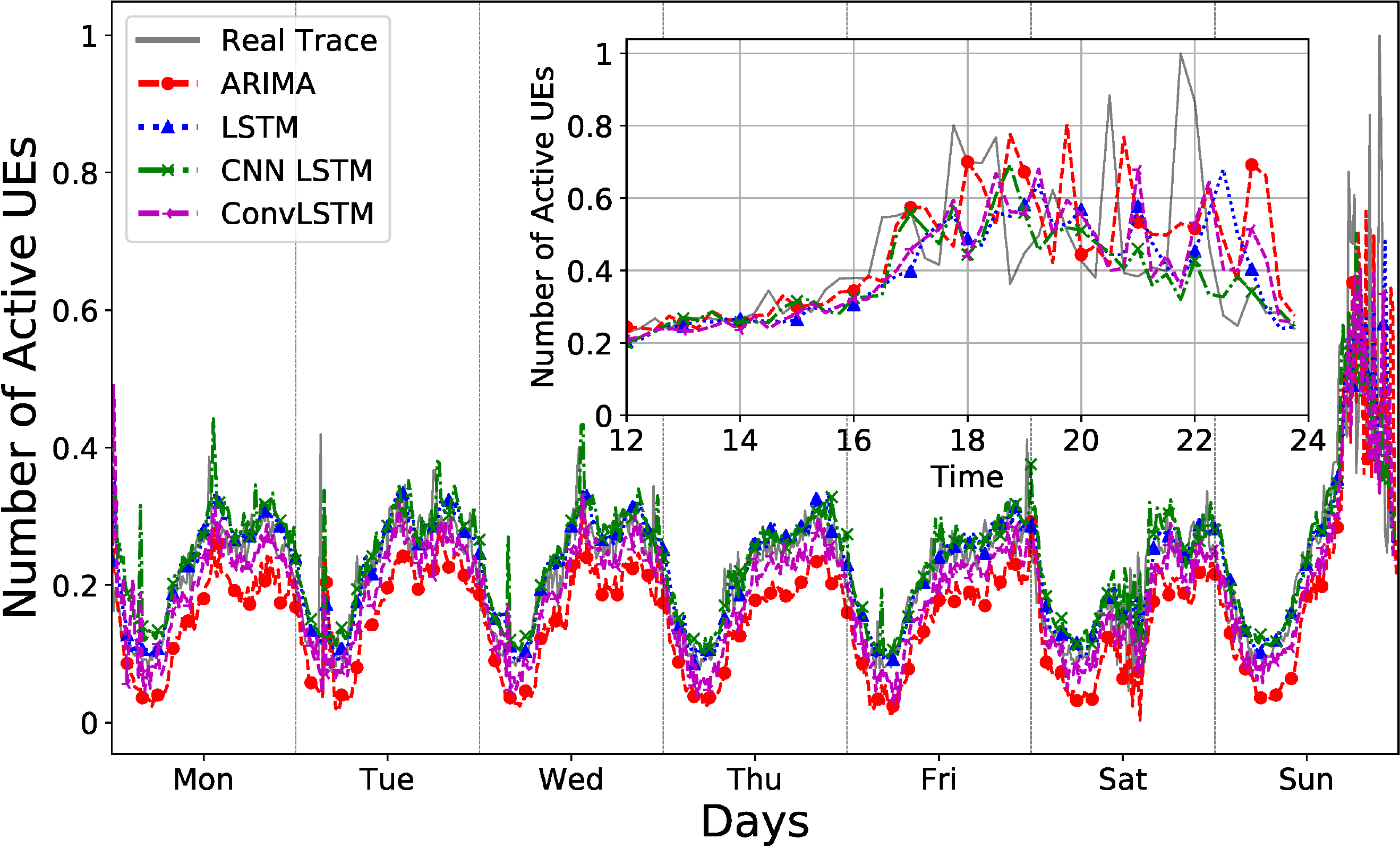}
\caption{\change{Comparison of different LSTM-based forecasting models working on real data from a base station in the stadium area (Normalized values).}}
\label{fig:LSTM_Comparison}
\end{figure}

\change{Fig.~\ref{fig:LSTM_Comparison} provides a comparative view of the models exploiting a walk-forward validation. Using such settings, the model is retrained as soon as new observations are made available thus expanding the time window horizon in each training step. Past predictions are then stored and evaluated against a longer list of observations, leading to more accurate forecasting performances.}
The different models exhibit good performance in forecasting the number of active users during normal working conditions thanks to the highly repetitive patterns. However, as highlighted by the inner-plot of Fig.~\ref{fig:LSTM_Comparison}, the prediction quality deteriorates during the event time period, resulting in an underestimation of the activity peaks. Due to its asymmetry with respect to standard working conditions, it is clear that temporal information alone is not enough to accurately predict the mobile traffic activity in the stadium during public events. In the following, we thus propose a deep learning model that in addition to spatio-temporal information also exploits contextual information to enhance the accuracy of the predictions.



\subsection{Resource utilization}
The relationship between QoS and availability of spectrum resources, characterized by functions $\zeta_u^e(c_b^e)$ and $q_b^e(c_b^e)$, is unknown and hard to construct due to the rarity and particularity of mass events. We hence focus on model-free deep learning methods to build our capacity forecasting mechanism. 

\begin{figure}[!t]
\centering
\includegraphics[width=\columnwidth, clip, trim = 0cm 0cm 0cm 0cm]{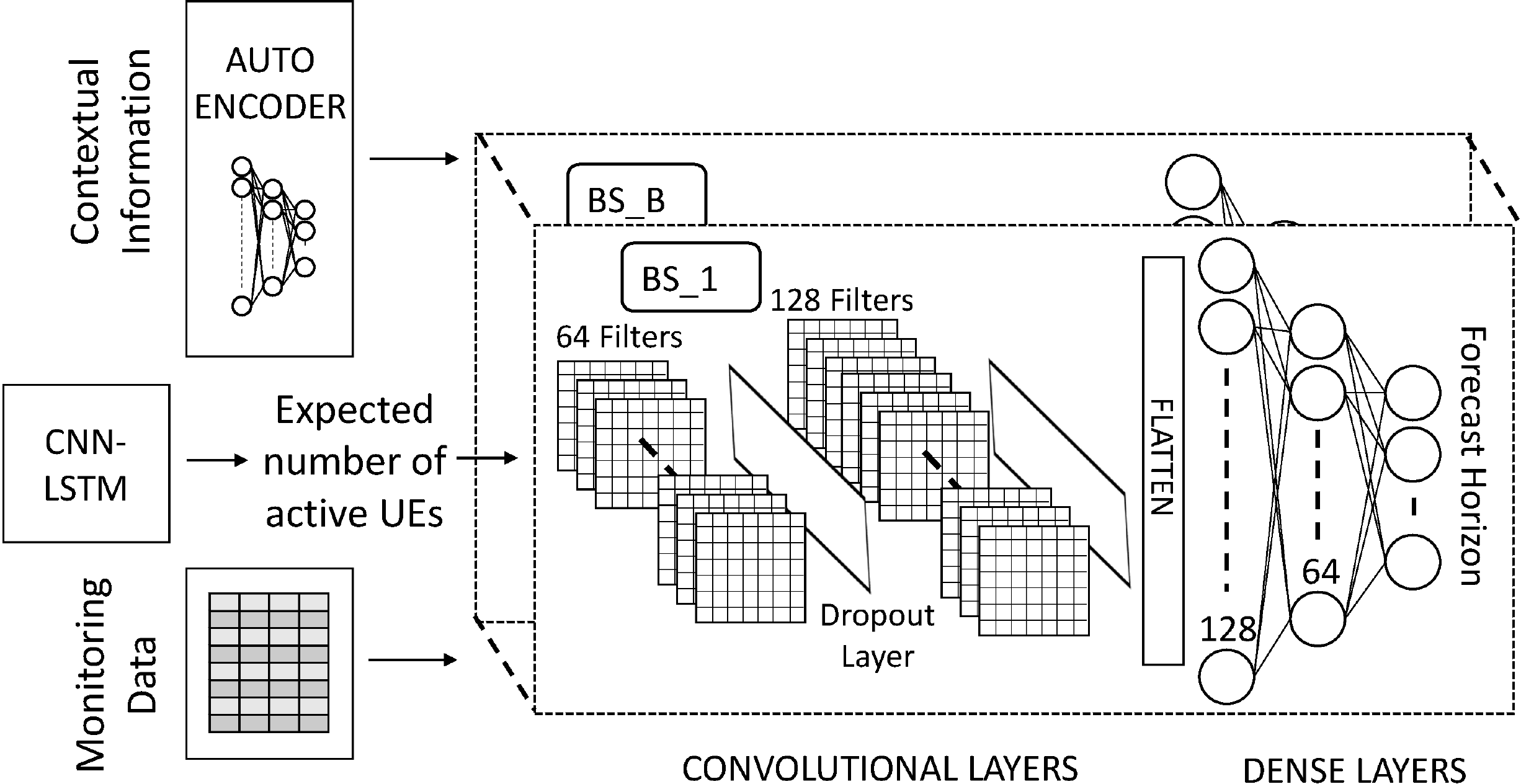}
\caption{Deep neural network architecture used in \name.}
\label{fig:cnn}
\end{figure}

To this aim, we design a neural network structure per base station, each receiving two sets of inputs pertaining each respective eNB: $i$) a representation of contextual information over the event (such as expected number of attendees, type of the event, etc.), and $ii$) network-specific monitoring data over time (historical time series), as depicted in Fig.~\ref{fig:cnn}. In order to reduce the dimensionality of the contextual data yet extract meaningful information we use a standard sparse autoencoder (SAE)~\cite[Ch.14]{Auxiliary_Information, Goodfellow-et-al-2016
}. In brief, a SAE consists of two feed-forward neural networks: an encoder (with an output layer of size $1$ in our case) and a decoder  (with an output layer of size equal to the dimensions of the input of the encoder). They are trained together so that the reconstructed output of the decoder is as \emph{similar} as possible to the input of the encoder. During exploitation, we simply use the encoder part of our SAE, which effectively compresses the contextual information into a 1-dimensional value. For the purpose of this study, we concentrate on the expected number of attendees to the event as contextual data, and empirically select the CNN-LSTM model as main forecasting module in light of the results discussed in Sec.~\ref{subsec:Active_users}.
On the other hand, the second set of inputs concerns the historical traces of network measurements introduced in~Section~\ref{sec:analysis}. In particular, we select the time evolution of average connected users $U_b^e$ (for base station $b$ during epoch $e$), volume of downlink traffic as well as our QoS metric $q^e_b$.
Thus, we train a deep neural network for each base station in the stadium area using $60\%$ of the football events in our dataset and validate the outcome using the remaining $40\%$.

Fig.~\ref{fig:cnn} depicts the overall architecture of our system, consisting in two neural networks connected in series. The first neural network accounts for two layers of two-dimensional Convolutional Neural Network (2D-CNN) with $\eta$ and $\mu$ filters respectively. The CNN is a class of deep neural network generally used in computer vision for their ability to detect patterns in input images. The workflow assumes each filter to be convolved along width and height of the input image to produce an activation map. Different filters detect distinct features so that a set of activation maps are passed to the next layer of the CNN.
We leverage on this and select from the overall measurements matrix $\bf M$ the information related to a specific base station $b$, i.e., $\bf M_b$. Then, we split $\bf M_b$ in epochs, or snapshots $s^e_b$, generating smaller input matrices of size $(F\times T)$, where $F$ is the number of features and $T$ is forecast horizon. We adopt a sliding window over time as data augmentation technique. A normalization function $\mathcal{N}(\cdot)$ is applied at this point to favour the learning process.

The neurons of the 2D-CNN apply a filter $\mathcal{H}(\sum_{e\in \mathcal{E}} \bf{I_e} \circledast \bf{K} + \bf{b_e} )$ where $\bf{I_e}$ is the input matrix at epoch $e$ (i.e., $\bf{I_e} = s^e_b$),  $\circledast$ indicates the 2D convolution operator, $\bf{K}$ is the filter kernel, $\bf{b}$ is a bias vector and $\mathcal{H}(\cdot)$ is a non linear activation function.
The significant development of deep learning methods lead to the definition of plethora of different activation functions in the literature~\cite{Activation_Comparison}. We empirically select the Rectified Linear Units (ReLu) function to overcome the vanishing gradient problem and allow the models to learn faster.
In order to reduce overfitting in favour of generalization, the inner layers of the CNN are interleaved with a Dropout layer. At each training stage, individual nodes and related incoming/outgoing edges are either dropped (with probability $p$) or kept (with probability $1-p$). This step allows decreasing the co-dependency from adjacent neurons during the training phase. The dropout probability is fixed to $0.2$.
The output of our CNN consists on a Flatten layer, so we map the input matrices into a fixed-length vector that feeds the second stage of the architecture.

\begin{figure}[t!]
\centering
\includegraphics[width=\columnwidth, clip, trim = 0cm 0cm 0cm 0cm]{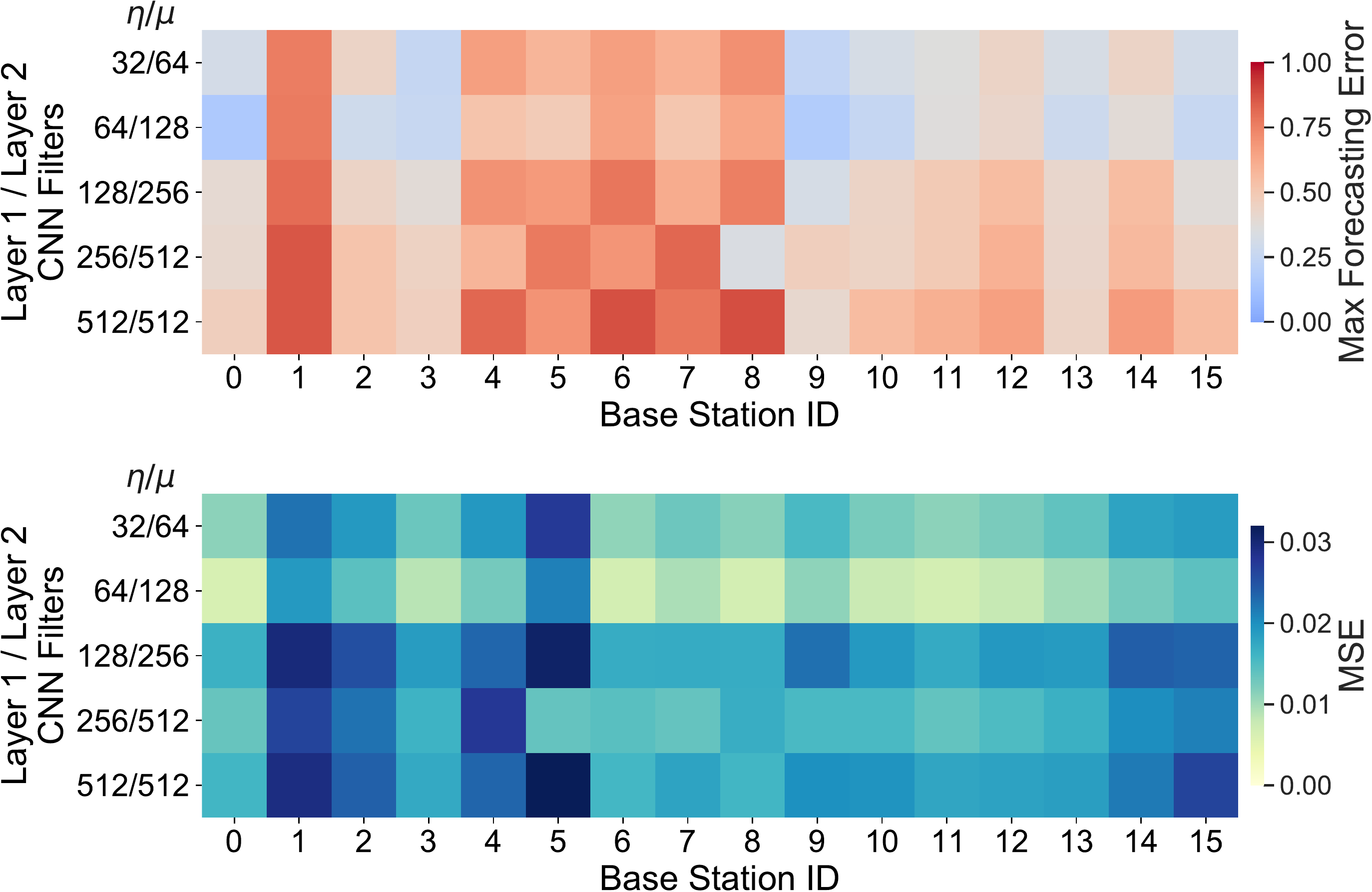}
\caption{Impact of different hyperparameter settings on \name{} forecasting performances.}
\label{fig:hyperparameter}
\end{figure}

\begin{figure*}[!t]
    \centering
	  \begin{subfigure}[b]{.32\textwidth}
          \centering
          \includegraphics[trim = 0cm 0cm 0cm 0cm,clip=false, width=5.5cm, keepaspectratio]{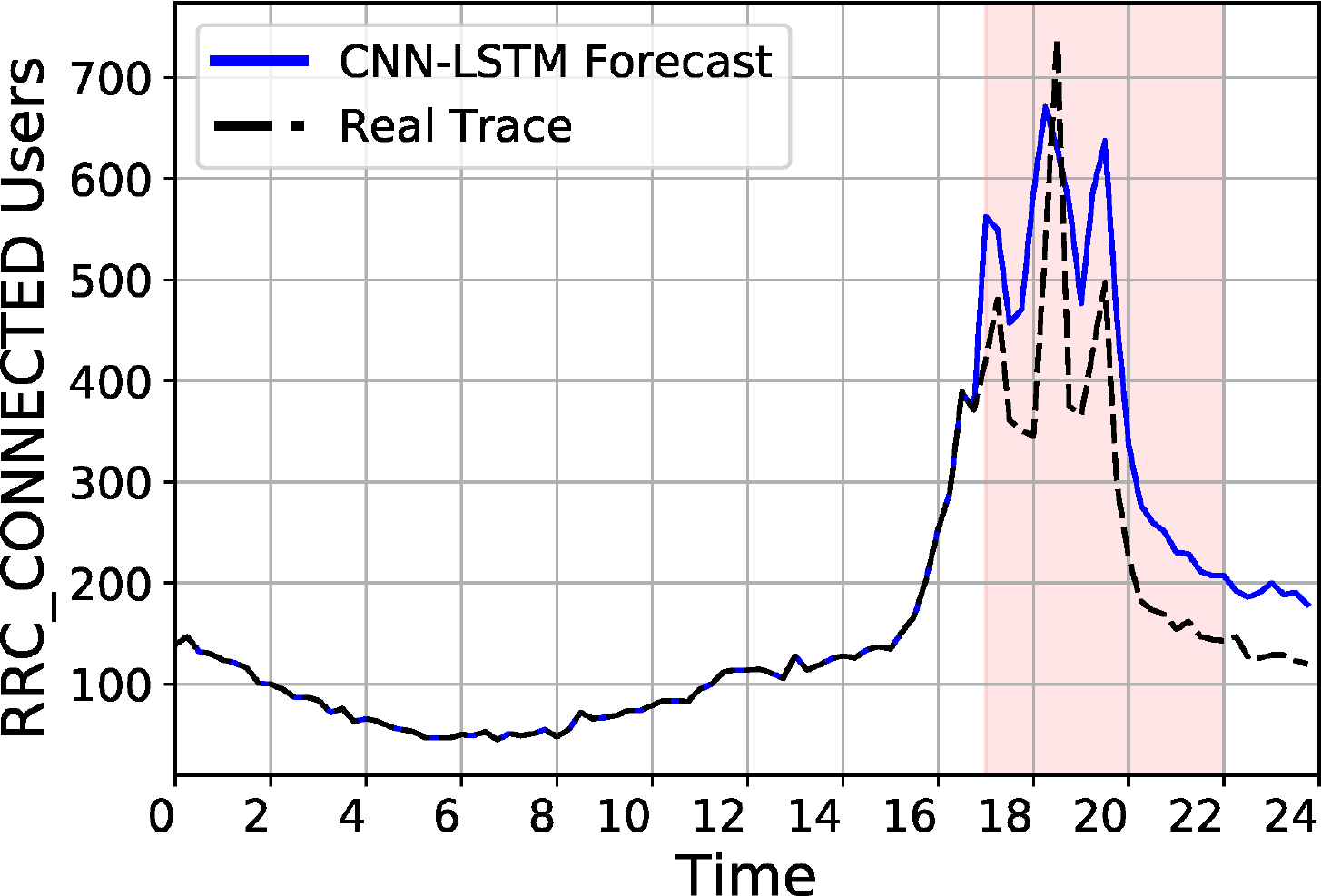}
          \caption{Sector 1}
          \label{fig:2600RRC}
      \end{subfigure}%
	  \begin{subfigure}[b]{.32\textwidth}
	        \centering
            \includegraphics[trim = 0cm 0cm 0cm 0cm,clip, width=5.5cm, keepaspectratio]{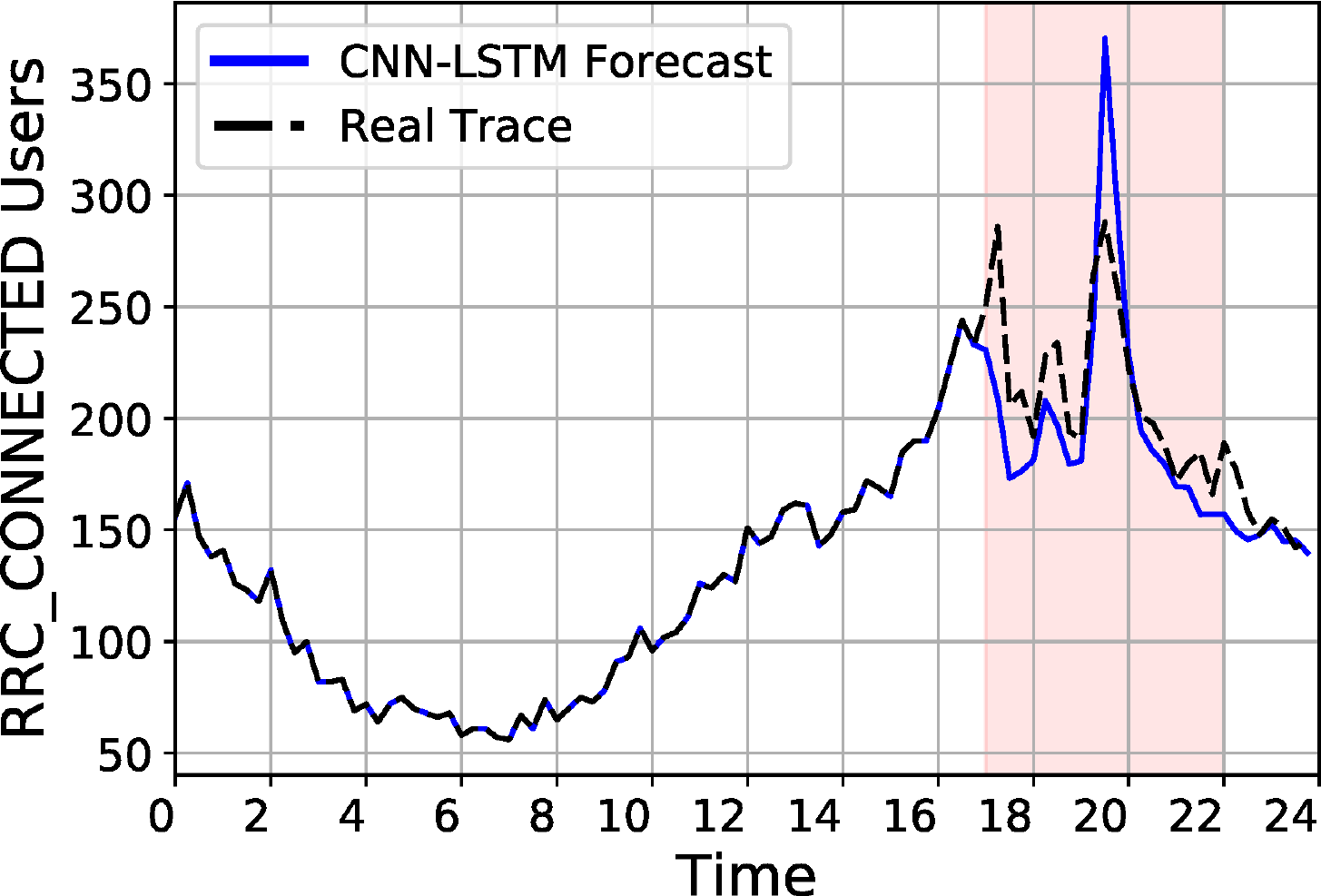}
            \caption{Sector 2}
            \label{fig:1800RRC}
      \end{subfigure}%
	  \begin{subfigure}[b]{.32\textwidth}
            \centering
            \includegraphics[trim = 0cm 0cm 0cm 0cm,clip, width=5.5cm, keepaspectratio]{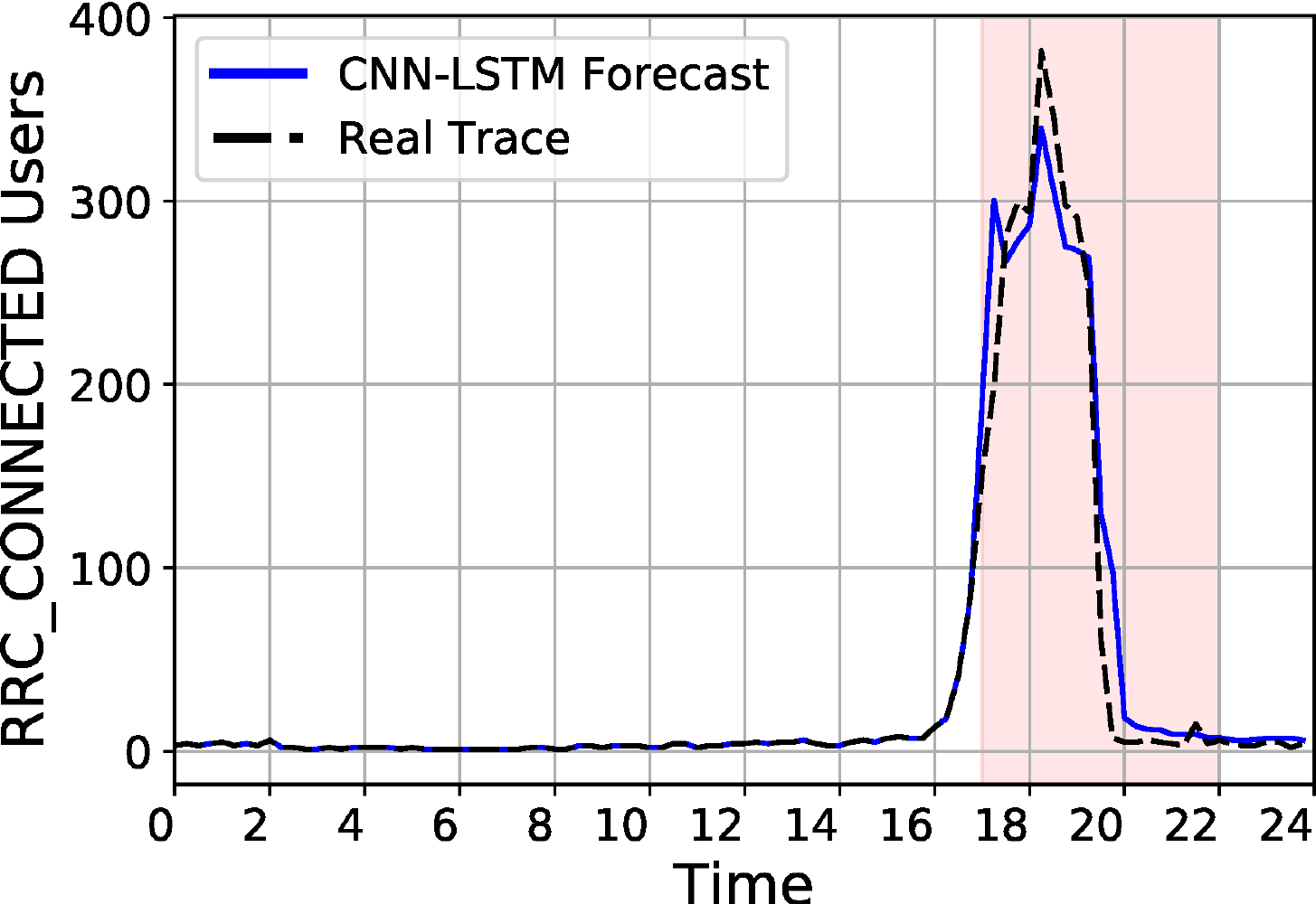}
            \caption{Sector 3}
            \label{fig:800RRC}
      \end{subfigure}%
      
      \vspace{4mm}
	  \begin{subfigure}[b]{.32\textwidth}
            \centering
            \includegraphics[trim = 0cm 0cm 0cm 0cm,clip, width=5.5cm, keepaspectratio]{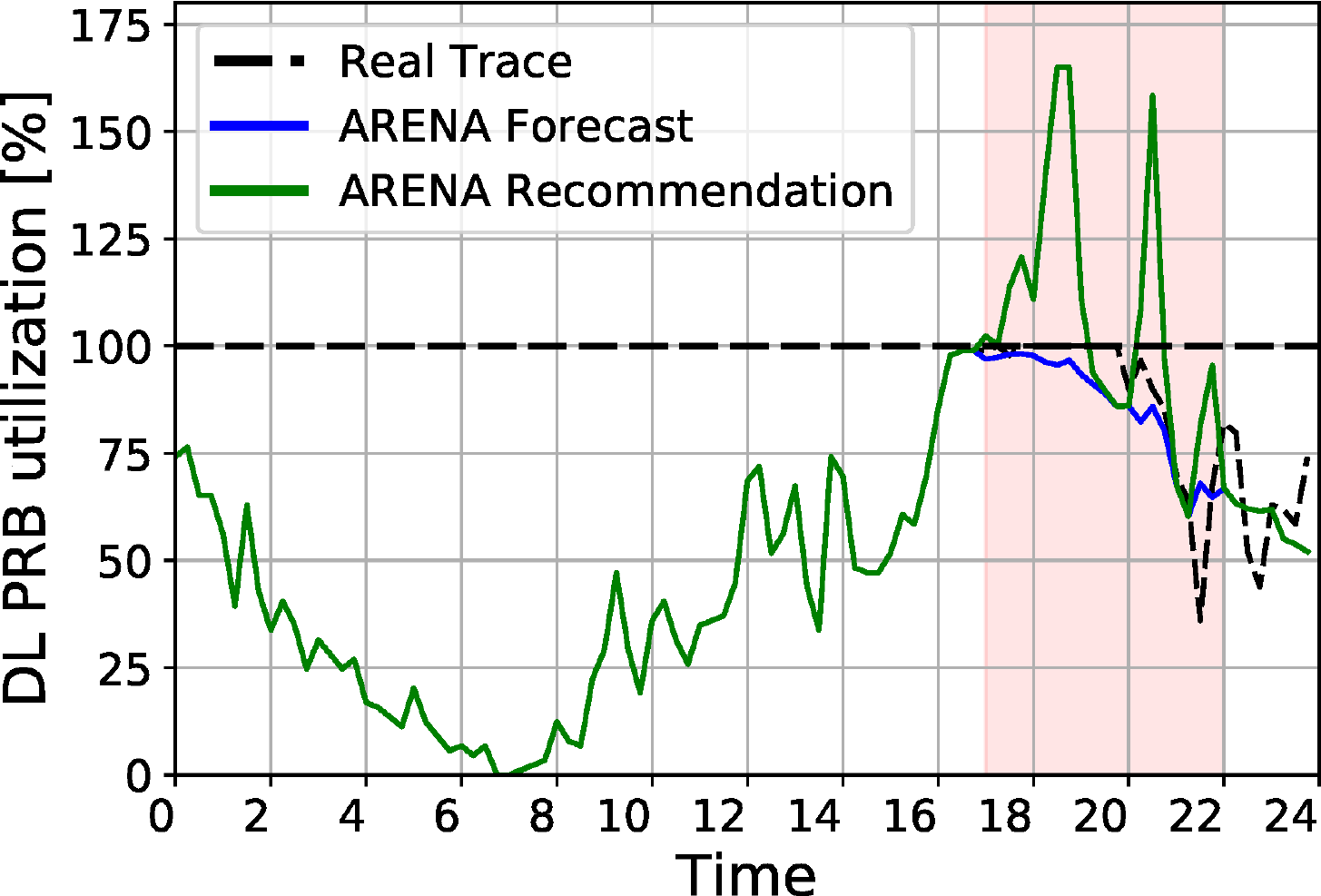}
            \caption{Sector 1}
            \label{fig:2600PRB}
      \end{subfigure}%
	  \begin{subfigure}[b]{.32\textwidth}
            \centering
            \includegraphics[trim = 0cm 0cm 0cm 0cm,clip, width=5.5cm, keepaspectratio]{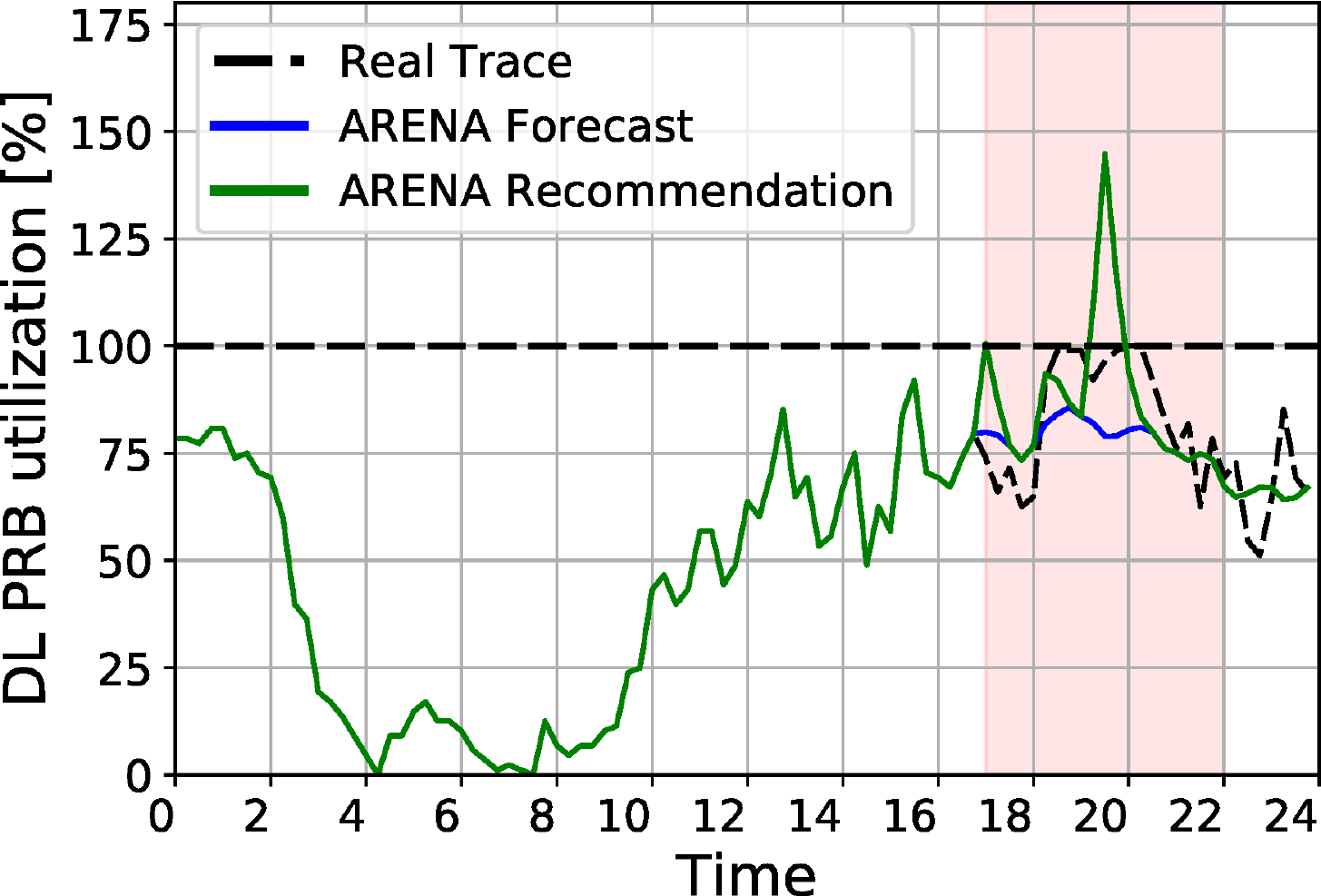}
            \caption{Sector 2}
            \label{fig:1800PRB}
      \end{subfigure}%
	  \begin{subfigure}[b]{.32\textwidth}
            \centering
            \includegraphics[trim = 0cm 0cm 0cm 0cm,clip, width=5.5cm, keepaspectratio]{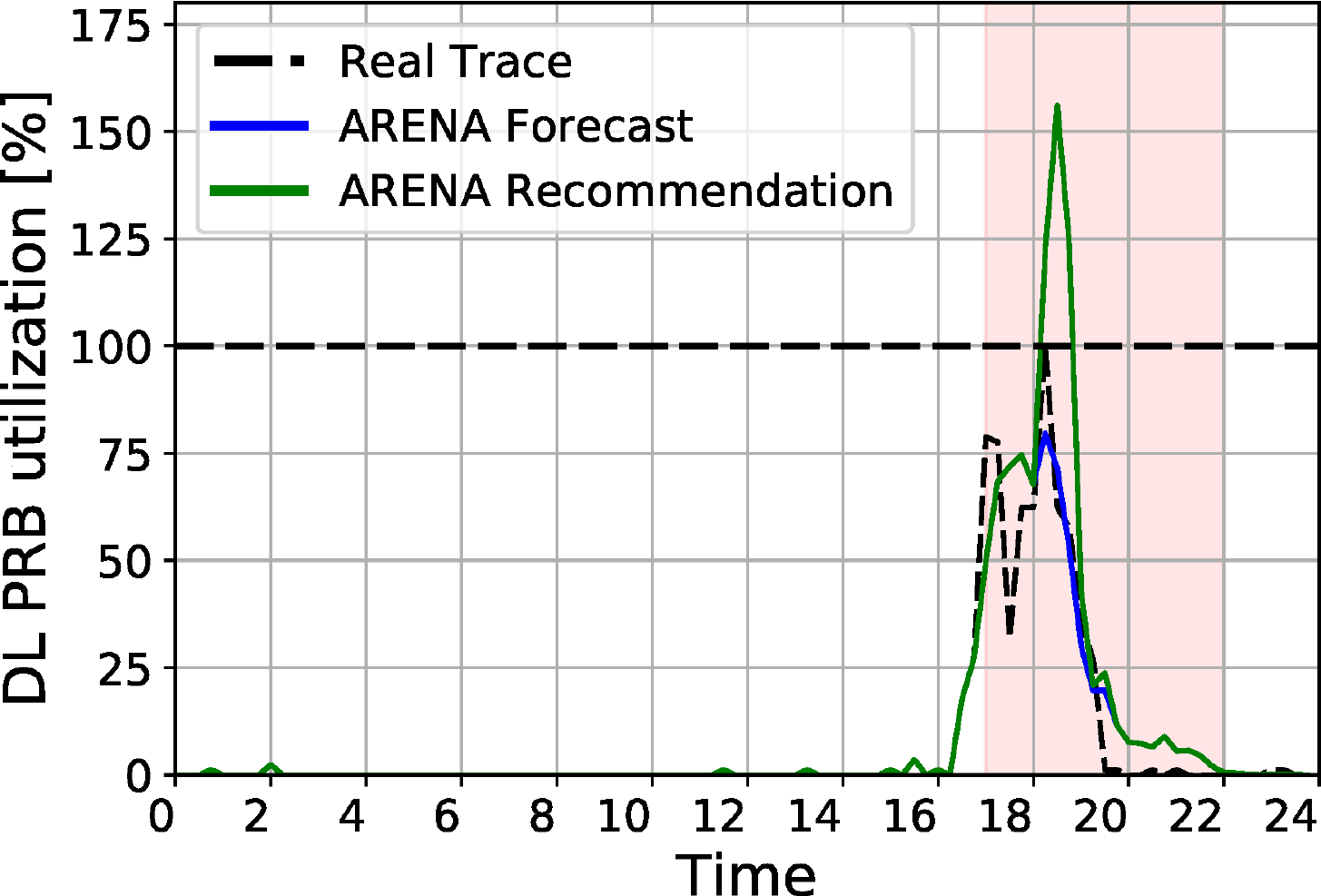}
            \caption{Sector 3}
            \label{fig:800PRB}
      \end{subfigure}%
 
    \caption{Forecast of the number of active users or $\texttt{RRC}_\texttt{CONNECTED}$  ($U_b$) during a mass event (blue line) and actual evolution (dashed black line) for three sectors chosen for illustration purposes. Forecast on the radio resource usage (blue straight line), actual usage evolution (dashed black line) and estimation of extra radio resource allocation (green line) for the same sectors.}
	\label{fig:sims_results_}
	\vspace{-5mm}

\end{figure*}

The second neural network consists in three feed-forward layers. In this structure, also called Multi-Layer Perceptrons (MLPs), the neurons of one layer are connected with all the neurons of the adjacent layer to build a finite and acyclic graph.
The number of neurons in the first and second layer are fixed to $128$ and $64$ respectively, while the size of the output layer of the network matches with the number of samples to be predicted $T$ for both output estimations: the active number of users and the PRB utilization during the event. Importantly, the MPLs inherit from standard multi-layer feed-forward networks the capability to approximate continuous $n$-dimensional functions $f:\mathbb{R}^n \to \mathbb{R}^n$ as stated by the universal approximation theorem~\cite{Approximation_Theorem}.
Thus, the output of each MPL layer is a linear combination of weighted real-valued input and a nonlinear activation function, namely $y = \phi(\bf{w}^{T}\bf{x} + \bf{b})$, where $\bf{w}$ denotes the vector of weights, $ \bf{x}$ is the vector of inputs, $\bf{b}$ is the bias and $\phi(\cdot)$ is the non-linear activation function.
Also in this case, we adopt the ReLu function for the hidden layers and a linear function for the output layer to retrieve the forecasting samples amplitude.
The output of the neural network includes a $T$-dimensional vector predicting the amount of PRB utilization $\bar{\vv{c}}_b$ during the $T$ epochs covering the event and a $T$-dimensional vector with the respective prediction on the number of active users $\bar{U}_b$.
The training phase exploits the adaptive moment estimation (Adam) optimizer, with learning rate $10^{-4}$. At each iteration, the model parameters are adjusted to minimize the error between predictions and ground truth measured by means of a loss function. In our model, we exploit the Mean Squared Error (MSE) loss function~\cite{Fiore_TMC18}.

\change{The choice of the best hyperparameter settings for \name{} has been performed by means of an accurate optimization process aiming at reducing the forecasting error. In Fig.\ref{fig:hyperparameter} we investigate the effects of a variable number of convolutional filters $\eta$ and $\mu$ over the two convolutional layers included in our proposed architecture. To deal with the highly heterogeneous traffic patterns, we have individually trained each model exclusively accounting for the traffic traces coming from the respective base station. The methodology to train each model follows the one previously presented. The training phase accounts for $50$ epochs and runs over $60\%$ of the overall dataset, while the validation phase encompasses the remaining $40\%$.
Our aim is to deal with mass events within the stadium premises over the event days, therefore, in the upper plot of Fig.\ref{fig:hyperparameter} we depict the maximum forecasting error experienced while estimating the resource consumption of each base station. From the picture, we can notice that for some base stations the resulting performances are poor, regardless the configuration settings of our model. This outcome can be easily explained as follows. Such considered base stations are generally underutilized in regular days. This leads the forecasting models to provide biased estimations of the cell capacity utilization, even in case of significant traffic loads generated during the sport events.

From the picture it can be noticed that an increase in the convolutional layer size yields a general performance degradation, both in terms of MSE and maximum forecasting error. Adding more layers may help to extract more features, but it also increases the number of parameters composing the model that, in turn, augments the chances of overfitted models.
Empirically, we have identified the best performance with $\eta=64$ and $\mu=128$ settings. Hence, we adopt these values throughout our paper.
}

With the above neural network structure, our system learns the relationship between active users and the incurred load. Our goal however is to provide additional spectrum capacity such that the same QoS experienced by the users during regular days is preserved during the event. To this aim, we let $\Delta^e_b=\frac{\bar{U}^e_b}{\tilde{U}_b}$ denote a scaling factor, where $\bar{U}^e_b$ is the predicted number of active users during epoch $e$ for base station $b$ (also marked as the output of our neural network) and $\tilde{U}_b$ is the number of users required to saturate resources of base station $b$ according to a radio usage behavior from a regular day---this can be learned from previous history. The motivation behind is to compensate the expected radio resource usage with respect to the predicted impact that extra active users might have on individual QoS during the event.
As a result, we finally devise the amount of spectrum required during the event as
\begin{align}
    &\vv{c}_b  = \bar{\vv{c}}_b\cdot\vv{\Delta}_b^T, &\forall b\in\mathcal{B}.\label{eq:arena}
\end{align}

\subsection{Performance evaluation}
\label{sec:evaluation}
Hereafter, we evaluate \name{} considering the real traffic traces collected within the stadium neighbourhood. We leverage the neural network capability to approximate the relationship between QoE and PRB utilization as well as the temporal correlation of the measurements (as described in Section \ref{sec:model}) in order to measure the amount of additional resources required to cope with the traffic demand during football events.
To this aim, we set the time horizon from which our neural network starts making predictions to $2$ hours before a real event and then we assess the ability of \name{} to predict the active number of users and spectrum usage after such time horizon. Based on this information, \name{} may take optimal spectrum allocation decisions. A complete snapshot of our validation results is depicted in Fig.~\ref{fig:sims_results_}.
On the one hand, Figs.~\ref{fig:sims_results_}(a)-(c) (top three plots) show the evolution over time of the active users for three different sectors chosen for illustration purposes. The black dashed line represents the actual data whereas the blue straight line shows the predicted values. Like before, the area in light red highlights the period of time when the event takes place. Note that both lines overlap before the time horizon of the event as they represent known information from the history. We observe that our neural network structure follows the time evolution of active users during the event remarkably well thereby capturing the particularities of the football match as we have discussed in our analysis in Section~\ref{sec:analysis}. 
On the other hand, Figs.~\ref{fig:sims_results_}(d)-(f) (bottom three plots) show the evolution over time of the downlink spectrum utilization for the same sectors. Similarly, the dashed line represents the actual time series whereas the straight blue line shows the predicted values spawned by our neural network structure. \change{In addition, we show the recommended PRB allocation with a straight green line (c.f. Eq.~\eqref{eq:arena}), obtained varying the input QoS metric as in standard operating conditions.}
As it can be observed, our prediction on spectrum usage follows closely the real usage until the radio access network capacity limit is reached. In such cases, the recommended PRB allocation value (green line)  follows the dynamics of the active users keeping the amount of time wherein extra capacity is required (PRB allocation above $100\%$) limited, thereby minimizing the incurred cost.




\section{Conclusions}
\label{sec:conclusions}

In this paper we performed a data analysis of the network dynamics during sport events in the neighbourhood of a football stadium. The mobile infrastructure covers a $1$-Km$^2$ area with $16$ base station sectors and is operated by a major European carrier. The events contained traffic peaks generated by about $30.000$ people.

Based on the insights obtained, we took up on the mass events RAN capacity forecasting challenge and designed \name{}: a model-free deep learning solution that predicts the expected number of connected users during a future event along with the corresponding RAN spectrum capacity (PRB) needed per sector for providing an average user experience. \name{}  takes as input the average number of connected users, downlink traffic volume, downlink throughput and other contextual event information to infer the future traffic loads.
Our results show that \name{} $i)$ closely predicts the actual number of connected devices during mass events in time and $ii)$ that it can provide mobile operators guidance on the actual RAN capacity needed during an event to provide a user experience similar to a regular network situation.



\bibliographystyle{IEEEtran}
{
\bibliography{main}
}

\end{document}